\documentclass[10pt,twocolumn,letterpaper]{article}

\usepackage[pagenumbers]{cvpr} % To force page numbers, e.g. for an arXiv version

\usepackage{graphicx}
\usepackage{amsmath}
\usepackage{amssymb}
\usepackage{booktabs}
\usepackage{algorithm}
\usepackage{algorithmic}
\usepackage[algo2e]{algorithm2e}
\usepackage{enumerate}
\usepackage{enumitem}
\usepackage{makecell}
\usepackage[numbers,sort&compress]{natbib}
\usepackage[usenames, dvipsnames]{color}
\usepackage{color}
\usepackage{subcaption,graphicx}
\usepackage[normalem]{ulem}
\usepackage[table]{xcolor}
\usepackage[usenames, dvipsnames]{color}
\usepackage{pifont}% http://ctan.org/pkg/pifont
\usepackage{color, colortbl}

\definecolor{Gray}{gray}{0.93}

\newcommand{\cmark}{\color{ForestGreen}\ding{51}}%
\newcommand{\xmark}{\color{red}\ding{55}}
\newcommand{\rulesep}{\unskip\ \vrule\ }

\usepackage[pagebackref,breaklinks,colorlinks]{hyperref}

\usepackage[capitalize]{cleveref}
\crefname{section}{Sec.}{Secs.}
\Crefname{section}{Section}{Sections}
\Crefname{table}{Table}{Tables}
\crefname{table}{Tab.}{Tabs.}

\def\cvprPaperID{1752} % *** Enter the CVPR Paper ID here
\def\confName{CVPR}
\def\confYear{2022}

\begin{document}
\setlength{\abovedisplayskip}{3pt}
\setlength{\belowdisplayskip}{3pt}
\title{EyePAD++: A Distillation-based approach for joint Eye Authentication and Presentation Attack Detection using Periocular Images}

\author{Prithviraj Dhar\textsuperscript{1}, Amit Kumar\textsuperscript{2}, Kirsten Kaplan \textsuperscript{2}, Khushi Gupta\textsuperscript{2}, Rakesh Ranjan\textsuperscript{2}, Rama Chellappa\textsuperscript{1}\\
\textsuperscript{1}Johns Hopkins University,
\textsuperscript{2}Reality Labs, Meta\\
{\tt\small \{pdhar1,rchella4\}@jhu.edu, \{akumar14,kkaplan,khushigupa,rakeshr\}@fb.com}
}
\maketitle

%%%%%%%%% ABSTRACT
\begin{abstract}
A practical eye authentication (EA) system targeted for edge devices needs to perform authentication and be robust to presentation attacks, all while remaining compute and latency efficient. However, existing eye-based frameworks a) perform authentication and Presentation Attack Detection (PAD) independently and b) involve significant pre-processing steps to extract the iris region. Here, we introduce a joint framework for EA and PAD using periocular images. While a deep Multitask Learning (MTL) network can perform both the tasks, MTL suffers from the forgetting effect since the training datasets for EA and PAD are disjoint. To overcome this, we propose Eye Authentication with PAD (EyePAD), a distillation-based method that trains a single network for EA and PAD while reducing the effect of forgetting. To further improve the EA performance, we introduce a novel approach called EyePAD++ that includes training an MTL network on both EA and PAD data, while distilling the `versatility' of the EyePAD network through an additional distillation step. Our proposed methods outperform the SOTA in PAD and obtain near-SOTA performance in eye-to-eye verification, without any pre-processing. We also demonstrate the efficacy of EyePAD and EyePAD++ in user-to-user verification with PAD across network backbones and image quality.
\end{abstract}

%%%%%%%%% BODY TEXT
\vspace{-0.3cm}
\section{Introduction}
Eye Authentication (EA) using irises has been widely used for biometric authentication. With the current advancements in head-mounted technology, eye-based authentication is likely to become an essential part of authenticating users against their wearable devices. While highly accurate, EA systems are also vulnerable to 'Presentation Attacks' (PA) \cite{he2016multi,yadav2019detecting}.  These attacks seek to fool the authentication system by presenting artificial eye images, such as printed iris images of an individual \cite{czajka2013database, hoffman2018convolutional}, or cosmetic contacts \cite{hughes2013detection}.  %While face recognition has also become one of the most preferred approaches for biometric authentication, it is not suitable to be used in AR/VR devices that only capture the periocular region of the face (and not the entire face).
\begin{figure}
\centering
{\includegraphics[width=\linewidth]{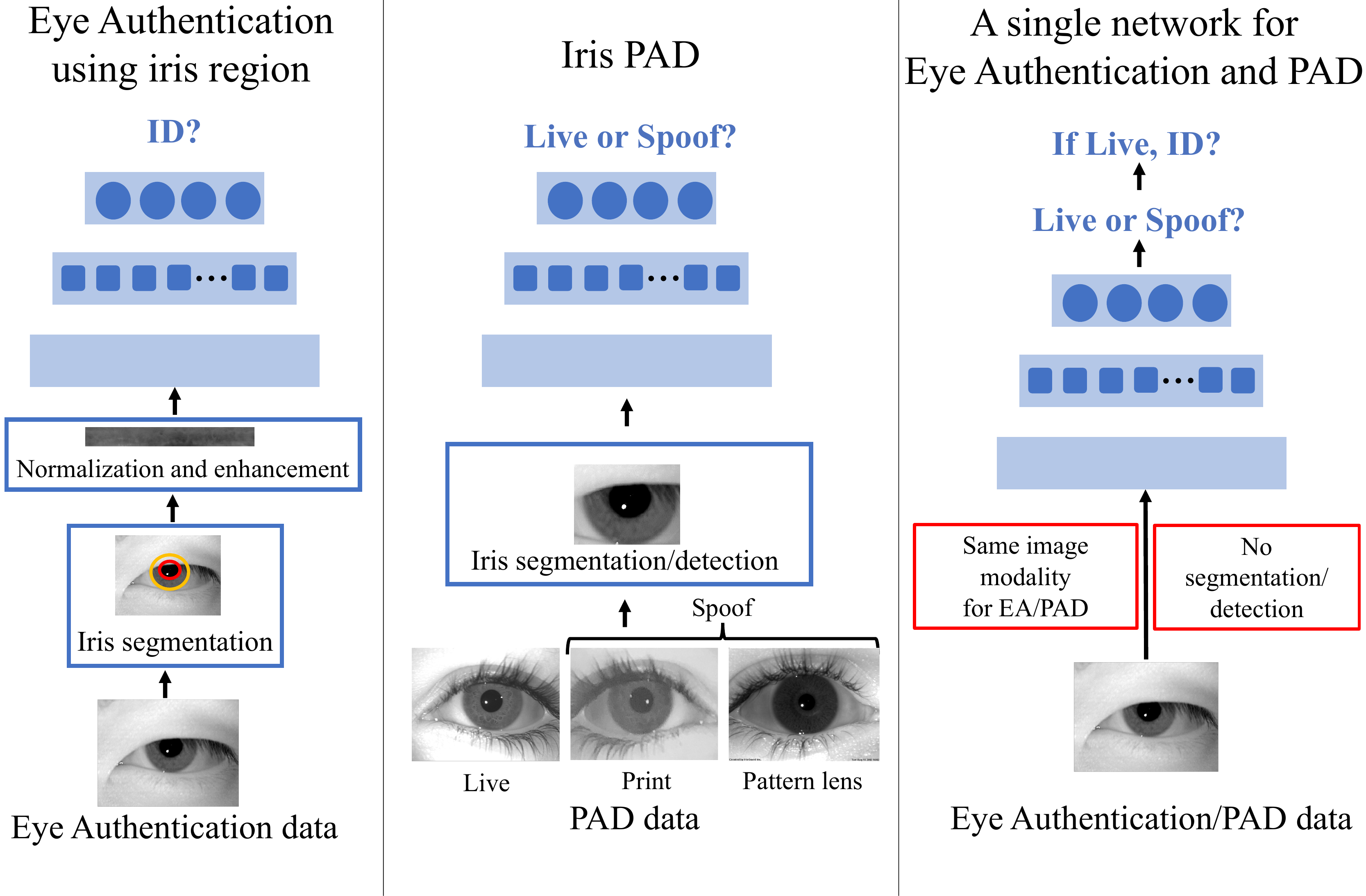}}
\vspace{-0.8cm}
\caption{\small EA pipelines segment out the iris region from the periocular image and normalize the iris image before feeding it to the network. Segmentation/detection is also used in most PAD pipelines. We train a single network for both EA and PAD without any pre-processing steps, using the entire periocular image.}\vspace{-0.5cm}
\label{fig:teaserapad}
\end{figure}
While researchers have proposed methods to train networks that achieve SOTA performance in either EA or PAD, a practical eye-based biometric system that can be used on edge devices must be able to perform both of these tasks accurately and simultaneously, with low latency and in an energy efficient manner. Therefore, in this work we propose strategies to develop a single deep network for both EA and PAD using periocular images.

One of the key steps in EA is pre-processing. Most EA frameworks \cite{zhao2017towards,wang2019toward} use an auxiliary segmentation network to extract the iris region from the periocular image. The iris is then unwrapped into a rectangular image and is fed to the eye authentication system. This geometric normalization step was first proposed in \cite{daugman1993high}. Pre-processing is also an important step in iris PAD pipelines that requires another segmentation network \cite{he2016multi,yadav2019detecting}, or a third party iris detection software \cite{rathgeb2016design,sharma2020d}. Such pre-processing steps potentially make the EA and PAD pipelines computationally expensive, making it impractical to embed these biometric systems on edge devices with limited computational resources. We investigate and propose techniques to perform EA and PAD using the entire periocular image without any active pre-processing. In doing so, we adhere to our goal of using a \textit{single} network in a truer sense.

For a given subject, irises of left and right eyes demonstrate different textural patterns. In most of the existing works in EA \cite{zhao2017towards,wang2019toward}, the CNN models are trained on the left irises for classification. During evaluation, a right iris is verified against the right irises of the same or other subjects. We refer to this evaluation method as `\textit{eye-to-eye verification}'.  However, a more practical protocol would be to perform user-to-user verification, i.e. consider both left and right eyes of a given test subject (query user) and verify it against one or more pairs of left-right irises of same or different user (i.e. gallery user). To this end, we propose a new  evaluation protocol to match the the left-right pair of a query user with that of a gallery user.

We consider the problem of EA and PAD as a disjoint multitask learning problem because the authentication task presumes real images, which is why the current datasets for EA do not include PAD labels. A possible single-network solution is to train a deep multitask network for both tasks, alternately training the EA and PAD branches with their respective dataset each iteration (as done in \cite{ranjan2017all}). However, several works \cite{li2017learning, kim2018disjoint, hong2020beyond} have shown that Multitask Learning (MTL) frameworks for disjoint tasks demonstrate forgetting effect (see Sec. \ref{sec:approach}). Hence, we propose two novel knowledge distillation-based techniques called EyePAD and EyePAD++ to incrementally learn EA and PAD tasks, while minimizing the forgetting effect. In summary, we make the following contributions in this work:
\begin{enumerate}[leftmargin=*]
    \item We propose a user-to-user verification protocol that can be used to authenticate one query user against one or many samples of a gallery user. This is more practical than the existing protocol for eye-to-eye verification.
    \item To the best of our knowledge, we are the first to explore the problem of EA and PAD using a single network. We introduce a new metric called Overall False Rejection Rate (OFRR) to evaluate the performance of the entire system (EA and PAD), using only authentication data.

    \item We propose a distillation-based method called \textbf{Eye} Authentication with \textbf{P}resentation \textbf{A}ttack \textbf{D}etection (EyePAD) for jointly performing  EA and PAD. To further improve the verification performance, we propose EyePAD++. EyePAD++ inherits the versatility of the EyePAD network through distillation and combines it with the specificity of multitask learning. EyePAD++ consistently outperforms the existing baselines for MTL, in terms of OFRR. We show the efficacy of EyePAD and EyePAD++ across different network backbones (Densenet121 \cite{huang2017densely}, MobilenetV3 \cite{howard2019searching} and HRnet64 \cite{wang2020deep}), and image quality degradation (blur and noise). Additionally, we apply our methods to jointly perform eye-to-eye verification and PAD, following the commonly used train-test protocols. Although the current SOTA approaches use pre-processing, our proposed methods outperform the existing SOTA in PAD task, and obtain comparable user-to-user verification performance without any pre-processing. 
\end{enumerate}
\begin{table}[]
\centering
%\scriptsize
\scalebox{0.75}{
\begin{tabular}{c|cc|ccccc}
\toprule
  Method  & EA & PAD &  Pre-processing \\
  \midrule
  IrisCode \cite{masek2003recognition}&\cmark &\xmark&Segmentation, geometric normalization\\
  Ordinal \cite{sun2008ordinal} &\cmark &\xmark&Segmentation, geometric normalization\\
  UniNet \cite{zhao2017towards}&\cmark &\xmark&Segmentation, geometric normalization\\
  DRFnet \cite{wang2019toward} &\cmark &\xmark&Segmentation, geometric normalization\\
  \cite{raghavendra2017contlensnet}&\xmark&\cmark&Segmentation, geometric normalization\\
  \cite{he2016multi}&\xmark&\cmark&Segmentation, geometric normalization\\
  \cite{pala2017iris}&\xmark&\cmark& Cropping\\
  DensePAD \cite{yadav2019detecting} &\xmark&\cmark&Segmentation, geometric normalization\\
\cite{hoffman2019iris+}&\xmark&\cmark&Segmentation with UIST \cite{rathgeb2016design}\\
  D-net-PAD \cite{sharma2020d} &\xmark&\cmark&Detection with VeriEye\\
  \cite{chen2021explainable} &\xmark&\cmark&Detection with \cite{chen2018multi}\\
  PBS, A-PBS \cite{fang2021iris} &\xmark&\cmark&None \\
  \midrule
  EyePAD (ours)& \cmark&\cmark & None\\
  EyePAD++ (ours) & \cmark&\cmark & None\\
 \midrule
\bottomrule
\end{tabular}
}
\vspace{-0.3cm}
\caption{\small Pre-processing steps in recent EA/PAD frameworks} \label{tab:relwork}
\vspace{-0.2cm}
\end{table}

\section{Related work}
\textbf{Eye authentication using irises:} Daugman \cite{daugman2009iris, daugman1993high} introduced the first automated system for EA by applying Gabor Filters to the normalized image for generating spatial barcode-like features (IrisCode). More recently, several works have proposed using deep features for EA. \cite{gangwar2016deepirisnet} proposed DeepIrisNet, the first deep learning-based framework for generalized EA, followed by \cite{he2017deep, nguyen2017iris, tang2017deep}. \cite{zhao2017towards} presents UniNet, that consists of two components: one for generating discriminative features (FeatNet) and the other for segmenting the iris and non-iris region (MaskNet). Both of these components accept the normalized iris images that also requires segmentation. \cite{wang2019toward} uses dilated convolution kernels for training CNNs for EA. \cite{yang2021dualsanet} presents an encoder-decoder pipeline to extract multi-level iris features and use an attention module to combine the multi-level features.\\
\begin{figure*}
{\centering
\subfloat[]{\includegraphics[width=0.45\linewidth]{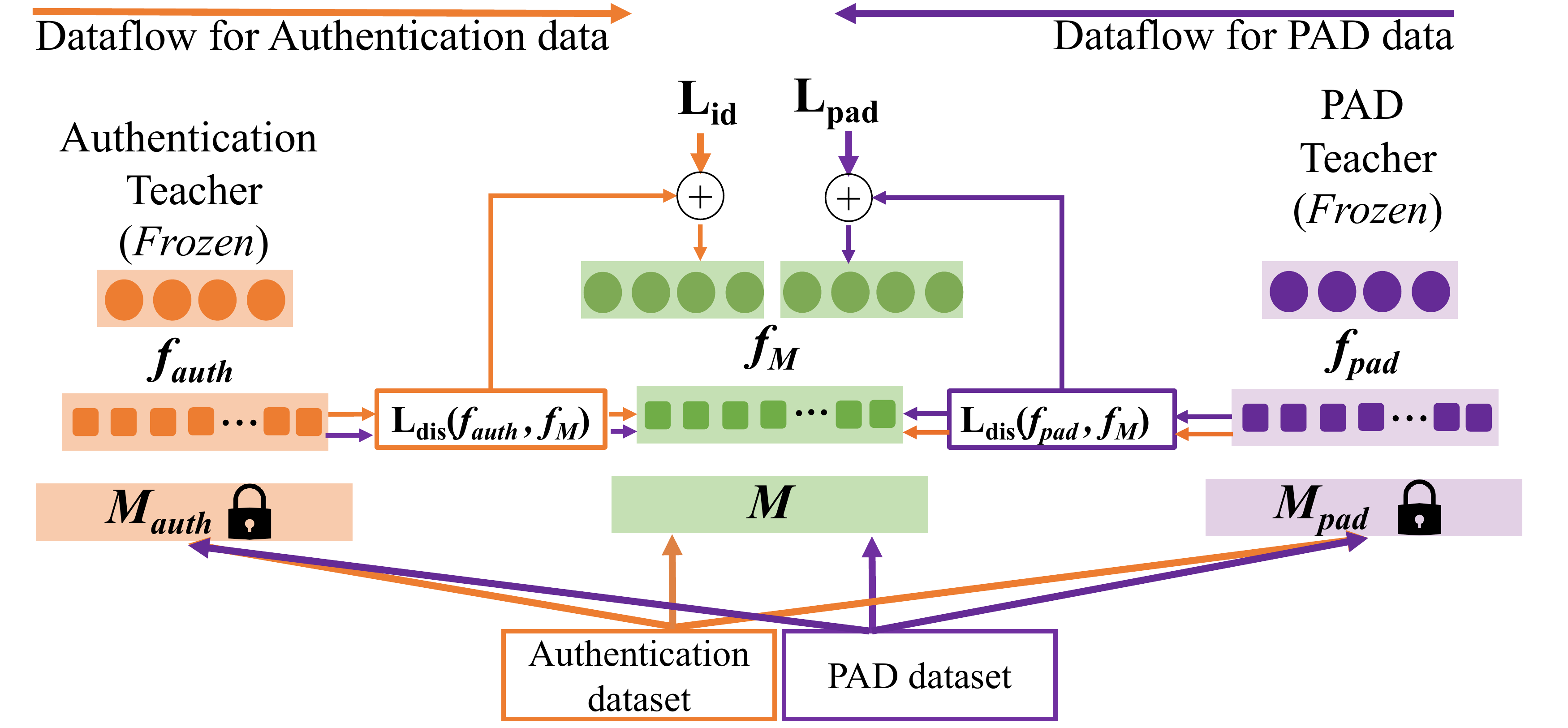}\label{fig:mtkd}}
\rulesep
\subfloat[]{\includegraphics[width=0.45\linewidth]{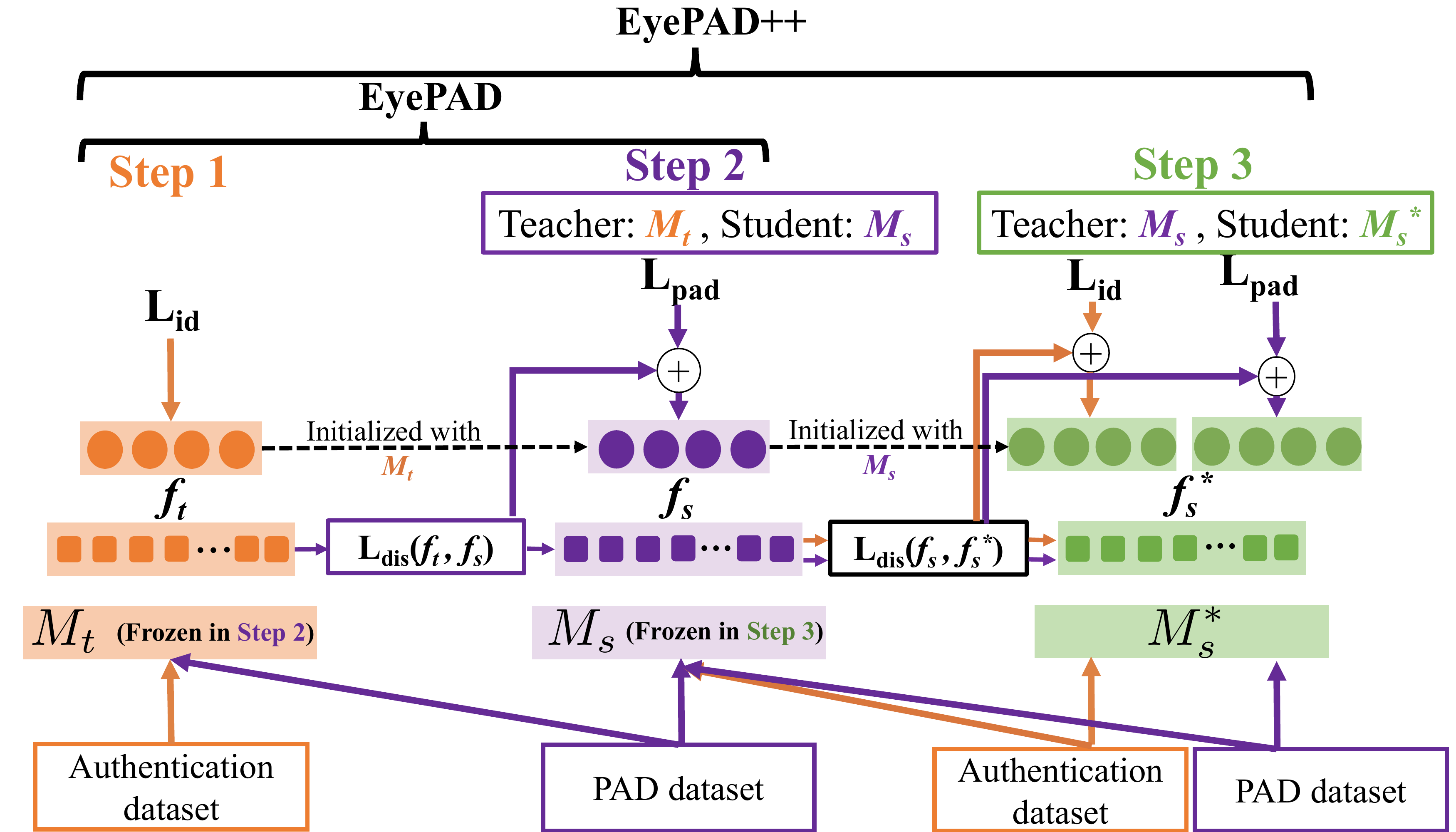}\label{fig:eyepad}}
\vspace{-0.3cm}
\caption{\small \textbf{(a)} \textbf{Baseline}: Multitask Learning with multi-teacher distillation (MTMT) \cite{li2020knowledge} \textbf{(b)} Proposed approach \textbf{Step 1}: We train $M_t$ for EA, \textbf{Step 2 (EyePAD)}: We initialize $M_s$ using $M_t$ and train it for PAD, while distilling EA information from $M_t$. (c) \textbf{Step 3 (EyePAD++)}: We initialize an MTL network $M^{*}_s$ with the trained $M_s$ and train it to perform both EA and PAD, while distilling the `versatility' of $M_s$. EyePAD++ outperforms the MTMT \cite{li2020knowledge} baseline in jointly performing EA and PAD in most of the problem settings.}
\vspace{-0.6cm}
}
\end{figure*}
\textbf{Eye-based Presentation Attack Detection:} PAD in periocular images has received significant attention from the deep learning community in the past few years \cite{pala2017iris,menotti2015deep,yadav2019detecting, sharma2020d,chen2021explainable, fang2021iris}. \cite{kuehlkamp2018ensemble, yadav2018fusion} propose fusing handcrafted and CNN features to detect PA. \cite{fang2020deep} fuses the features from different layers in a deep network extracted for normalized iris images for PAD. \cite{yadav2019detecting, sharma2020d} show that DenseNet architecture helps to achieve high PAD accuracy. \cite{hoffman2019iris+} proposes dividing the iris region into overlapping patches and training CNNs using these patches. \cite{chen2021explainable} introduces an attention guided mechanism to improve PAD accuracy. \cite{fang2021iris} introduces a binary pixel-wise supervision with self attention to help the network to find patch-based cues and achieve high performance in PAD.\\
All of the EA algorithms use the normalization process proposed in \cite{daugman1993high} that requires iris segmentation. Similarly, most PAD algorithms also use auxiliary pre-processing steps such as iris detection/segmentation. A brief summary of the preprocessing steps in EA and PAD is given in Table \ref{tab:relwork}.\\
\textbf{Disjoint Multitask Learning and Knowledge Distillation:} 
Disjoint multitask learning (MTL) is the process of training a network to perform multiple tasks using data samples that have labels for either of the tasks, but not for all the tasks. Training a single network for EA and PAD is a disjoint MTL task because EA datasets do not include PAD labels. One solution is to follow the existing disjoint multitask learning strategies \cite{ranjan2017all, kokkinos2017ubernet, liu2016hierarchical} and update each branch of the network alternately. However, it is well known \cite{li2017learning, kim2018disjoint} that alternating training suffers from the forgetting effect \cite{li2017learning} and degrades performance in multitask learning. Knowledge Distillation (KD) \cite{44873} has been commonly used to reduce forgetting in continual learning \cite{ li2017learning, rebuffi2017icarl, romero2014fitnets, dhar2019learning,zagoruyko2016paying}. Inspired by this, \cite{kim2018disjoint, li2020knowledge} employ feature-level KD for multitasking. In \cite{kim2018disjoint}, KD is used to distill the information from the network from a previous iteration $i-1$ that was updated for task A (teacher), while training it to perform task B in the current iteration $i$ (student). However, in this scenario, the teacher network is not fully trained in the initial few iterations and thus the distillation step may not help preserve task A information. Similar to \cite{kim2018disjoint}, we propose strategies employing feature-level KD for disjoint multitasking (EA and PAD). But, unlike \cite{kim2018disjoint}, we ensure that the teacher network in our proposed methods is fully trained in one or more tasks.
\vspace{-0.3cm}
\section{Proposed approach}
\label{sec:approach}
Our objective is to build a single network that is proficient in performing two disjoint tasks: EA and PAD. We intend to build this framework for edge devices with limited on-device compute. Thus, we exclude any pre-processing step for detecting or segmenting the iris region and use the entire periocular image as input. Mutitask Learning (MTL) is a possible approach in this scenario. Most of the MTL methods for disjoint tasks \cite{ranjan2017all} alternately feed the data from different tasks. However, as shown in \cite{kim2018disjoint}, MTL demonstrates the forgetting effect. Consider an MTL network with shared backbone and different heads designed to perform two tasks A and B. Suppose that the training batches for task A and B are fed to this MTL network alternately. Here, the weights of the shared backbone modified by the gradients corresponding to the loss for task A in iteration $i$, may be rewritten in the next iteration ($i+1$) by the gradients corresponding to the loss for task B. This may lead to forgetting of task A.%Suppose we feed the batch for task A (say, EA) in a given iteration $i$ to the shared parameters and EA branch of an MTL network, and backpropagate the error through the EA branch and shared parameters of the network. Let's say, in the next iteration $i+1$,  we feed the next batch that corresponds to task B (say, PAD) to the shared parameters and PAD branch of the network. Now, after backpropagating the PAD error, the parameters of the shared backbone may get modified in a way that the previous updates corresponding to the EA might get forgotten. 
Therefore, instead of MTL, we propose to use knowledge distillation to learn both tasks through a single network. Here, we intend to first train a teacher network $M_t$ for EA, following which we train a student network $M_s$ for PAD, while distilling the authentication information from $M_t$ to $M_s$ to minimize the forgetting effect. %Incrementally learning PAD with distillation from $M_t$ helps to minimize the effect of forgetting, while simultaneously enabling the student $M_s$ learn both the tasks.
\subsection{Eye Authentication with Presentation Attack Detection (EyePAD) and EyePAD++}
\label{subsec:apad}
\setlength{\parindent}{0cm}{
We now explain the steps in our proposed methods: EyePAD and EyePAD++ (Fig. \ref{fig:eyepad}): } \\
\textbf{Step 1}: We train the teacher network $M_t$ using periocular images from the EA dataset to perform EA. Similar to \cite{wang2019toward}, we use triplet loss to train $M_t$. We first extract features $f_i$ for all the images using the penultimate layer of $M_t$. To select the $n^{th}$ triplet in a given batch, we randomly select an anchor feature $f^{(n)}_a$ belonging to category $C$ . After that, we select the hardest positive feature $f^{(n)}_{pos}$ and the hardest negative feature $f^{(n)}_{neg}$ as follows:  $$f^{(n)}_{pos} = \underset{i\in C, i \neq a}{\text{argmax}}(\|f^{(n)}_i- f^{(n)}_a\|^2), f^{(n)}_{neg} = \underset{i\not\in C}{\text{argmin}}(\|f^{(n)}_i-f^{(n)}_a\|^2)$$
Then we compute the triplet loss $L_{id}$ for the entire batch (of size $N$)  as:
\begin{equation}
    L_{id} = \frac{1}{N} \sum^{n=N}_{n=1} \text{max}(\|f^{(n)}_{pos} - f^{(n)}_a\|^2 - \|f^{(n)}_{neg} - f^{(n)}_a\|^2 + \alpha, 0)
    \label{eq:trplt}
    \vspace{-0.3cm}
\end{equation}
where $\alpha$ denotes the distance margin.\\
\textbf{Step 2 (Feature-level knowledge distillation - EyePAD)}: We initialize a student network $M_s$ using $M_t$, and train it for PAD. Let $I$ be an image from the PAD dataset. $I$ is fed to both $M_t$ and $M_s$, to obtain features $f_t$ and $f_s$, extracted using the penultimate layer of the corresponding networks. To constrain $M_s$ to process an eye image like $M_t$, we employ feature-level KD and minimize the cosine distance between $f_t$ and $f_s$ using the proposed distillation loss $L_{dis}$.
\begin{equation}
    L_{dis}(f_s,f_t) =1 - \frac{f_s\cdot f_t}{\|f_s\|\|f_t\|}.
\end{equation}
Our application of feature-level KD is inspired by \cite{romero2014fitnets}. Note that we do not apply KD on the output scores as done in \cite{li2017learning}, since eye-based matching protocols like \cite{wang2019toward} use the features from the penultimate layer (and not the output score vector). Additionally, we would like $M_s$ to classify a given image as live (also referred to as `real' or `bona-fide') or spoof using $L_{pad}$, which is a standard cross-entropy classification loss. Combining these constraints, we train $M_s$ using the multitask classification loss $L_{multi}$ as
\begin{equation}
    L_{multi} = L_{pad} + \lambda_1 L_{dis},
\end{equation}
where $\lambda_1$ is used to weight $L_{dis}$. In this step, the teacher $M_t$ remains frozen. To evaluate the verification performance, we use features extracted from the penultimate layer of the trained $M_s$ for the test EA data and perform user-to-user verification. We feed the test PAD data to $M_s$ and evaluate its performance in live/spoof classification. We find that the student network $M_s$ obtained from EyePAD is a versatile network that is effective for both EA and PAD. However, compared to $M_t$ (that was only trained for EA), $M_s$ obtains slightly lower verification performance. We hypothesize that $M_s$ demonstrates this drop in performance because it was never trained for EA. Therefore, we introduce an additional step to train an MTL network (initialized with $M_s$) while distilling the versatility of the EyePAD student to this network (Fig. \ref{fig:eyepad}).\\
\textbf{Step 3 (EyePAD++):} We initialize a new student network $M^{*}_s$ using $M_s$. $M^{*}_s$ is trained for both EA and PAD in an MTL fashion. Following the commonly used strategies for disjoint multitasking \cite{ranjan2017all}, the batches from EA and PAD data are alternated after every iteration. To reduce forgetting, we additionally constrain $M^{*}_s$ to mimic $M_s$, which acts as its teacher, using the same knowledge distillation used in step 2. We feed the training image to both $M_s$ and $M^{*}_s$ and obtain features $f_s$ and $f^{*}_s$ respectively. $M_s$ remains frozen in this step. We use them to compute $L_{dis}$ as: 
\begin{equation}
    L_{dis}(f_s,f^{*}_s) =1 - \frac{f_s\cdot f^{*}_s}{\|f_s\|\|f^{*}_s\|}.
\end{equation}
When authentication data is fed to $M^{*}_s$ (say, during iteration $i$), we compute $L^{id}_{multi}$ as follows:
\begin{equation}
    L^{id}_{multi} = L_{id} + \lambda_2 L_{dis}
\end{equation}
Here, $L_{id}$ is the triplet loss from Eq. \ref{eq:trplt}. When PAD data is fed to $M^{*}_s$ (during iteration $i+1$), we compute $L^{pad}_{multi}$ as:
\begin{equation}
    L^{pad}_{multi} = L_{pad} + \lambda_2 L_{dis}
\end{equation}
where $L_{pad}$ is a standard classification loss used in step 2 of EyePAD. Thus, $L^{id}_{multi}$ and $L^{pad}_{multi}$ are alternately used to optimize $M^{*}_s$. For inference, we use the trained $M^{*}_s$ for user-to-user verification and PAD. Hyperparameter details for EyePAD and EyePAD++ are provided in the supplementary material.
\section{Experiments}
\begin{figure}
\label{fig:tradeoffsupp}
\centering
\subfloat[]{\includegraphics[width=0.25\linewidth]{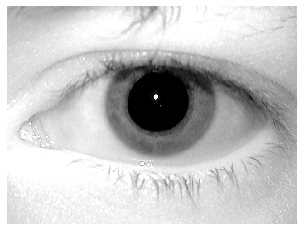}\label{fig:clean}}
~\subfloat[]{\includegraphics[width=0.25\linewidth]{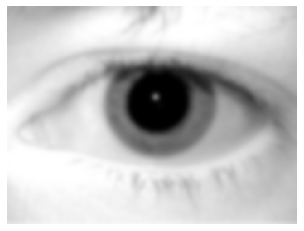}\label{fig:blur}}
\subfloat[]{\includegraphics[width=0.25\linewidth]{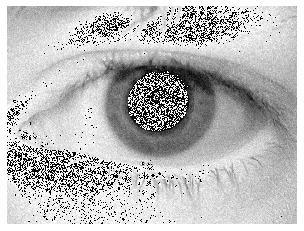}\label{fig:noise}}
\label{fig:qualdeg}
\vspace{-0.3cm}
\caption{\small We perform train and test our networks on (a) the original (clean) datasets, (b) their blurred and (c) their noisy versions. }
\vspace{-0.5cm}
\end{figure}
\subsection{Baseline methods}
\label{subsec:baseline}
\textbf{Single task networks}: To estimate the standard user-to-user verification (EA) and PAD performance, we test the `EA only' and `PAD only' networks. The `EA only' network is the teacher network $M_t$ used in Step 1 of EyePAD. \\
\textbf{Multitask Learning (MTL)}:  We train a multitask network for EA and PAD by alternately feeding  EA and PAD batches (see step 3 of EyePAD++) and alternately optimizing using $L_{pad}$ (Sec. \ref{subsec:apad}) and $L_{id}$ (Eq. \ref{eq:trplt}).\\
\textbf{Multi-teacher Multitasking (MTMT) \cite{li2020knowledge}}: MTMT \cite{li2020knowledge} is a recently proposed multitask framework that  combines MTL with multi-teacher knowledge distillation. MTMT has been shown to outperform MTL and other SOTA multitask methods such as GradNorm \cite{chen2018gradnorm}. Here, single task networks are first trained in specific tasks. An MTL network $M$ is then trained for multiple tasks, while information from the single task networks is distilled into $M$. A key difference between MTMT and EyePAD++ is that MTMT enforces distillation from multiple task-specific teachers whereas EyePAD++ includes distillation from a single teacher that is proficient in performing multiple tasks.  We implement MTMT as one of our baselines for joint EA and PAD (Fig. \ref{fig:mtkd}). Firstly, we train two single task models: $M_{auth}$ for EA and $M_{pad}$ for PAD, and then distill information from them while training a student MTL network $M$. We use the same feature-level distillation used in EyePAD (Step 2) and EyePAD++.  A given image is fed to $M_{auth}$, $M$, and $M_{pad}$, generating features from the penultimate layers $f_{auth}$, $f_{M}$ and $f_{pad}$, respectively. $L_{dis}$ then constrains $f_{M}$ to be closer to $f_{auth}$ and $f_{pad}$. $M_{auth}$ and $M_{pad}$ remain frozen in this step. We alternately feed the training batches for EA and PAD. So, when EA data is forwarded to $M_{auth}$, $M_{pad}$, $M$, we optimize $M$ using $L^{id}_{mtmt}$.
\begin{equation}
    L^{id}_{mtmt} = L_{id}+\lambda_{auth}L_{dis}(f_{auth},f_M)+\lambda_{pad}L_{dis}(f_{pad},f_M)
\end{equation}
Here $L_{id}$ is the triplet loss defined in Eq.\ref{eq:trplt}. $\lambda_{auth}, \lambda_{pad}$ denote the distillation weights from teacher $M_{auth}$ and $M_{pad}$, respectively. Similarly, when PAD data is forwarded, we optimize $M$ using $L^{pad}_{mtmt}$.
\begin{equation}
    L^{pad}_{mtmt} = L_{pad}+\lambda_{auth}L_{dis}(f_{auth},f_M)+\lambda_{pad}L_{dis}(f_{pad},f_M)
\end{equation}
Here $L_{pad}$ is standard classification loss for live/spoof classification. We provide the hyperparameter information for MTMT \cite{li2020knowledge} in the supplementary material. 
\begin{table}[]
\centering
%\scriptsize
\scalebox{0.75}{
\hskip-0.2cm\begin{tabular}{c|cc|cc}
\toprule
    & \multicolumn{2}{c|}{User-to-user verification (EA)}& \multicolumn{2}{c}{PAD}\\
    \midrule
    & Data & \# images &  Data & \# images  \\
  \midrule
  Train & \thead{206 users from\\ ND-Iris-0405}& 7949& \thead{Train split of\\ CU-LivDet (2013,2015,2017),\\ ND-LivDet (2013,2015,2017)} & 14600\\
  \midrule
  Test & \thead{150 users from \\ND-Iris-0405}& \thead{4231\\(2925 query,\\ 1306 gallery)}&\thead{Test split of CU-LivDet \\(2013,2015,2017)}&7532\\
\bottomrule
\end{tabular}
}
\vspace{-0.3cm}
\caption{\small Statistics for datasets used for EA with PAD} \label{tab:data}
\vspace{-0.4cm}
\end{table}
\subsection{Datasets and network architectures used}
\label{subsec:datarch}
We summarize the datasets used in our work in Table \ref{tab:data}.

\textbf{EA dataset}: We use the ND-Iris-0405 \cite{bowyer2016nd,phillips2009frvt} dataset, used widely for eye authentication using irises. The dataset consists of 356 users that are divided into two subsets: $U_{train}$ (randomly selected 206 users) and $U_{test}$ (remaining 150 users). Using the distinct left and right eye images for the users in $U_{train}$ gives us 412 (206$\times$2) categories, which we use to train models for EA. For a given user $u$ in $U_{test}$, we select 10 left and 10 right eye images to build the query set $q_u$. Similarly, we select 5 left and 5 right eye images to build the gallery set $g_u$ for user $u$. Repeating this for all the users in $U_{test}$, we obtain the query set $Q = \{q_u, \forall u \in U_{test} \}$ and gallery set $G = \{g_u, \forall u \in U_{test} \}$. To enable other researchers replicate our experiments, we provide the train and test splits in the supplementary material.\\
\textbf{PAD dataset}: For PAD training data, we combine the official training splits of CU-LivDet and ND-LivDet from the LivDet challenges in 2013\cite{6996283}, 2015\cite{7947701}, and 2017\cite{yambay2017livdet}. We build the PAD test dataset by combining the official test splits of the CU LivDet dataset from the 2013, 2015 and 2017 challenges. CU-LivDet consists of three categories: Live, patterned lens and printed images. ND-LivDet consists of two categories: Live and patterned lens. \\
\textbf{Image quality degradation}: The datasets we use in this work are academic datasets \cite{bowyer2016nd,6996283,7947701,yambay2017livdet} with high quality images (Fig. \ref{fig:clean}). However, real-world authentication on edge devices rely on small sensors which capture low-resolution images. Also, environmental conditions like lighting may further degrade the image quality. Therefore, in addition to using the original datasets, we also perform experiments by degrading the datasets (separately): 
(i) Blur: We add Gaussian blur with a random kernel size between 1 and 5 to the training images, and add blur with kernel size of 5 to the test images (Fig. \ref{fig:blur}). 
(ii) Noise: We add Additive White Gaussian Noise with a standard deviation $\sigma=3.0$ (Fig. \ref{fig:noise}).\\
\textbf{Networks used}: We implement our proposed methods and baselines using the Densenet121 backbone \cite{huang2017densely}. This is motivated by this architecture repeatedly demonstrating high PAD performance \cite{sharma2020d,fang2021iris,yadav2019detecting}. To demonstrate the generalizability of EyePAD and EyePAD++, we repeat our experiments using the HRnet64 \cite{wang2020deep} and MobilenetV3 \cite{howard2019searching}.
\vspace{-0.225cm}
\subsection{User to user verification protocol}
\begin{figure}
\centering
{\includegraphics[width=\linewidth]{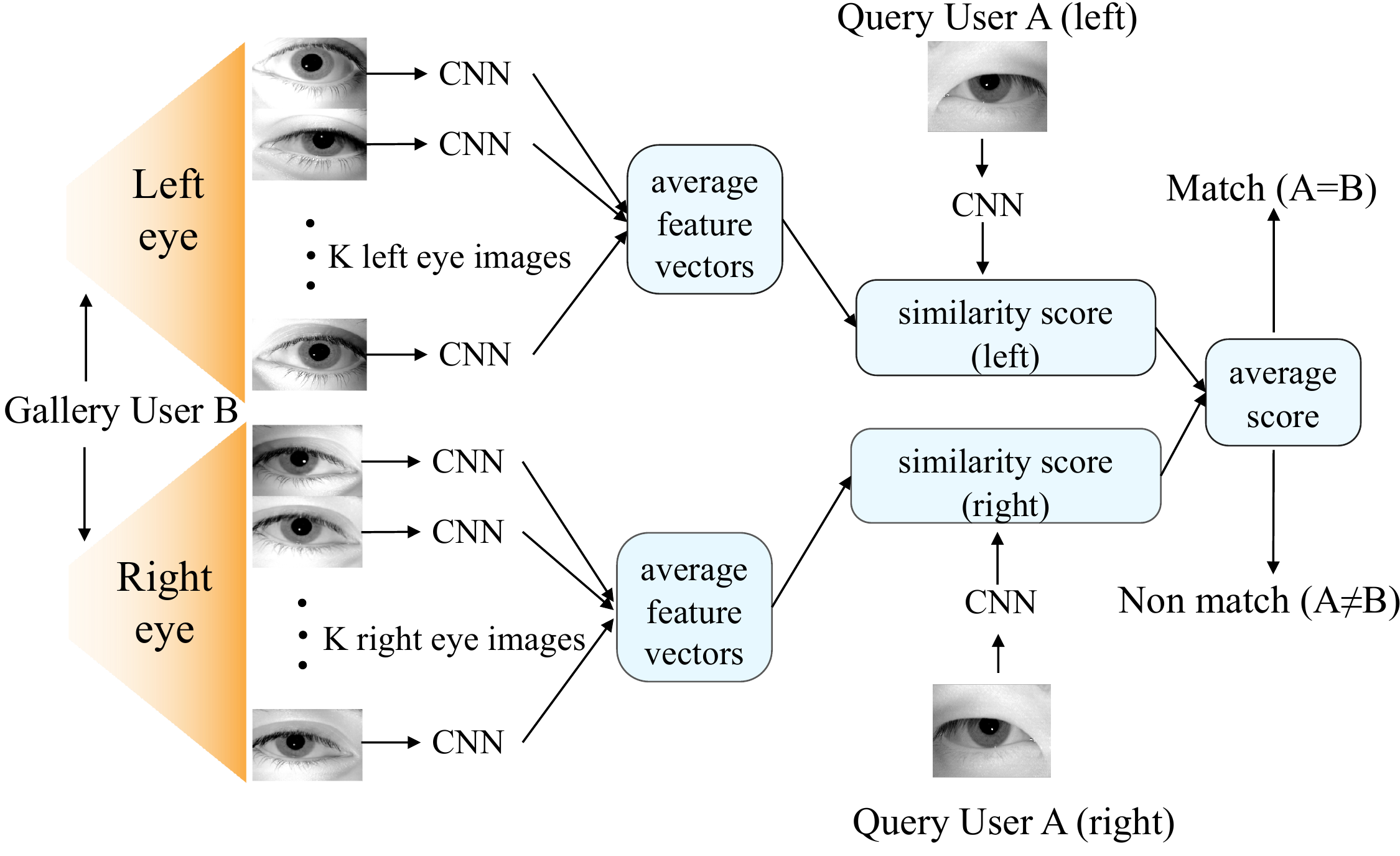}}
\vspace{-0.8cm}
\caption{\small User-to-user verification: Verifying query user A ($q_A$) against gallery user B ($g_B$) with $K$ pairs for gallery user B. }\vspace{-0.2cm}
\label{fig:u2u}
\end{figure}

 Most experiments in EA \cite{zhao2017towards,wang2019toward} train the model on the left irises of all the users and evaluate them in terms of the eye-to-eye verification accuracy for the right irises. However, in a real-world authentication system, the gallery will most likely have both left and right eye images (instead of only right eye images) for an authorized user, and thus both left and right query images can be used for verification. Moreover, it is more practical to authenticate a user using both eyes, as opposed to only the right eye. Hence, we propose matching one pair of left-right eyes (query) to $K$ pairs of left-right eyes (gallery).
\begin{algorithm}[t]
\algsetup{linenosize=\small}
 \small
\floatname{algorithm}{Protocol}
\caption{User to User verification (1 Query, $K$ Gallery)}
\label{alg:pass}
\begin{algorithmic}[1]
\STATE \textbf{Required}: Model $M$, Query dataset $Q$, Gallery dataset $G$
\STATE \textbf{Initialize}: Similarity dictionary $S$=[]
\STATE \textbf{Initialize}: Left and right Query dictionary $Q_L, Q_R$
\FOR { Query user $q_A$ in $Q$ and Gallery user $g_{B}$ in $G$}
\STATE Left query $q^{(L)}_A \leftarrow \text{RandomSelect}(q_A , 1, \text{Left})$
\STATE Right query $q^{(R)}_A \leftarrow \text{RandomSelect}(q_A , 1, \text{Right})$
\STATE $g^{(L)}_{B,1},g^{(L)}_{B,2} \ldots g^{(L)}_{B,K} \leftarrow \text{RandomSelect}(g_B , K, \text{Left})$
\STATE $g^{(R)}_{B,1},g^{(R)}_{B,2} \ldots g^{(R)}_{B,K} \leftarrow \text{RandomSelect}(g_B , K, \text{Right})$
\STATE $Q_L[q_A] = q^{(L)}_A $, $Q_R[q_A] = q^{(R)}_A $ 
\STATE Left query feature $f^{(L)}_{q_A} = M(q^{(L)}_A)$
\STATE Right query feature $f^{(R)}_{q_A} = M(q^{(R)}_A)$
\STATE Left and right gallery features $f^{(L)}_{g_B}, f^{(R)}_{g_B} = \frac{1}{K} \sum^{k=K}_{k=1} M(g^{(L)}_{B,k}), \frac{1}{K}\sum^{k=K}_{k=1} M(g^{(R)}_{B,k})$
\STATE Compute similarity $s(q_A,g_B)=\frac{1}{2}(\text{Similarity}(f^{(L)}_{q_A}, f^{(L)}_{g_B}) + \text{Similarity}(f^{(R)}_{q_A}, f^{(R)}_{g_B}))$
\STATE $S[q_A,g_B]\longleftarrow s(q_A,g_B)$
\ENDFOR
\STATE TAR, FAR, threshold = ROC($S$, EA Ground Truth)
\STATE $t_{auth}$ = threshold at FAR=$10^{-3}$
\end{algorithmic}
\label{prot:u2u}
\end{algorithm}
We provide the detailed user-to-user verification protocol in Protocol \ref{prot:u2u}. To match query user A $(q_A)$ and gallery user B $(g_B)$, we first randomly select one left eye and one right eye image from $q_A$. Then, we select $K$ left eye and $K$ right eye images from $g_B$. After that, we feed the left and right query images to a model $M$ and compute their respective features $f^{(L)}_{q_A}, f^{(R)}_{q_A}$. For gallery user B, we compute the features for the $K$ left eye images using $M$ and average them to compute a single feature $f^{(L)}_{g_B}$ (Line 12 of Protocol \ref{prot:u2u}). In the same way, we compute the average feature  $f^{(R)}_{g_B}$ by for the right eye gallery image. We then compute the similarity between query user $A$ and gallery user $B$ as :
\begin{equation}
    s(q_A,g_B) =\frac{1}{2} \Bigg( \frac{f^{(L)}_{g_B}\cdot f^{(L)}_{q_A}}{\|f^{(L)}_{g_B}\|\|f^{(L)}_{q_A}\|} + \frac{f^{(R)}_{g_B}\cdot f^{(R)}_{q_A}}{\|f^{(R)}_{g_B}\|\|f^{(R)}_{q_A}\|} \Bigg)
    \label{eq:sim}
\end{equation}
Based on the similarity threshold, a match/non-match is predicted (Fig. \ref{fig:u2u}). 
\begin{algorithm}[t]
\algsetup{linenosize=\small}
 \small
\floatname{algorithm}{Protocol}
\caption{Computing OFRR}
\label{alg:ofrr}
\begin{algorithmic}[1]
\STATE \textbf{Required}: Model $M$, Query dataset $Q$, Gallery dataset $G$
\STATE \textbf{Required}: Similarity dictionary $S$ from Protocol \ref{prot:u2u}
\STATE \textbf{Required}: Query dictionaries $Q_L, Q_R$ from Protocol \ref{prot:u2u}
\STATE \textbf{Required}: Similarity threshold $t_{auth}$ from Protocol \ref{prot:u2u}
\STATE \textbf{Required}: PAD threshold $t_{pad}$ for SAR=5\%
\STATE \textbf{Initialize}: Spoof rejects $X_{spoof}=0$, EA rejects $X_{auth}=0$
\FOR {Query user $q_A$, gallery user $g_A$ in $Q,G$}
\STATE $q^{(L)}_A \leftarrow Q_L[q_A]$, $q^{(R)}_A \leftarrow Q_R[q_A]$
\STATE Left query PAD logit $o^{(L)}_{q_A} = M(q^{(L)}_A)$
\STATE Right query PAD logit $o^{(R)}_{q_A} = M(q^{(R)}_A)$
\IF{$o^{(L)}_{q_A} > t_{pad}$ or $o^{(R)}_{q_A} > t_{pad}$}
\STATE $X_{spoof}=X_{spoof}+1$ // falsely rejected as spoof
\ELSE\IF{$S[q_A,g_A] < t_{auth}$}
\STATE $X_{auth}=X_{auth}+1$ // falsely rejected as non-match
\ENDIF
\ENDIF
\ENDFOR
\STATE Overall false rejection rate OFRR = $(X_{spoof} + X_{auth})/|Q|$
\end{algorithmic}
\label{prot:ofrr}
\end{algorithm}
\begin{table*}[]
\centering
%\scriptsize
\scalebox{0.75}{
\hskip-0.7cm\begin{tabular}{c|cccc|cccc|cccc|cccc}
\toprule
     & \multicolumn{12}{c|}{User-to-user verification results on ND-Iris-0405 (EA)} & \multicolumn{4}{c}{PAD results on CU-LivDet}\\
 \midrule
 & \multicolumn{4}{c|}{1 Query 1 Gallery}&  \multicolumn{4}{c|}{1 Query 2 Gallery}&  \multicolumn{4}{c|}{1 Query 5 Gallery}&\\
 \midrule
  Method & OFRR$(\downarrow)$ & $10^{-4}$ &  $10^{-3}$&$10^{-2}$&  OFRR$(\downarrow)$ &$10^{-4}$ &  $10^{-3}$& $10^{-2}$&  OFRR$(\downarrow)$ &$10^{-4}$ &  $10^{-3}$ &$10^{-2}$&   TDR$(\uparrow)$ & APCER &BPCER& HTER$(\downarrow)$ \\
  \midrule
  %0.88618148 0.95786631 0.99577492
  EA only& - & 0.886 & 0.958 & 0.996 & -& 0.891 & 0.954 & 0.994 &- & 0.921 & 0.979& 0.997 &- & -& -&-\\
  PAD only & -&-&-&-&-&-&-&-&-&-&-&-& 0.971 & 0.026 & 0.003 & 0.015\\
  MTL & 0.100& 0.693 & 0.905 &0.988&   0.074&0.803& 0.943 &0.986& \underline{0.052} & 0.850 &0.956 &0.990& 0.950 & 0.036&0.004&0.020\\
  MTMT \cite{li2020knowledge} &0.087& 0.919 &0.963& 0.992& 0.068 & 0.872& 0.950& 0.989 & \underline{0.052} & 0.816& 0.923 & 0.986&0.945&0.051&0.003&0.027\\
  \rowcolor{Gray}
  EyePAD &  \underline{0.079}& 0.843 &0.926 &0.993&  \textbf{0.046}& 0.887&0.961& 0.997 &0.060 & 0.922&0.952& 0.995&0.947& 0.036 & 0.014 & 0.025\\
  \rowcolor{Gray}
  EyePAD++ & \textbf{0.072}&0.901& 0.952& 0.990 & \underline{0.055}& 0.906& 0.966 & 0.997&  \textbf{0.043} &0.929& 0.983& 0.996 &0.951& 0.034& 0.012&0.023\\
  \midrule
  EA only & -& 0.832 &0.916 & 0.979 &-& 0.867 & 0.943 & 0.985 &-& 0.909&  0.966& 0.992 &-&-&-&-\\
  PAD only& -&-&-&-&-&-&-&-&-&-&-&-& 0.844 & 0.073 & 0.032 & 0.053\\
  MTL& 0.244& 0.745 &0.871&0.974& 0.209& 0.801 &0.910&0.989& 0.199& 0.834& 0.930& 0.994&0.702&0.136&0.034&0.085\\
  MTMT \cite{li2020knowledge}& 0.236 & 0.753 & 0.894 & 0.974 & 0.213 & 0.841 & 0.921 & 0.986& \textbf{0.189} & 0.822 & 0.946 & 0.993 & 0.576 & 0.047 & 0.095 &0.071\\
  \rowcolor{Gray}
  EyePAD & \underline{0.231}&  0.757&0.876&0.979 & \underline{0.204}& 0.796& 0.933& 0.974 & 0.196& 0.813& 0.949 & 0.987& 0.738 & 0.129&0.024 & 0.077\\
  \rowcolor{Gray}
  EyePAD++ &\textbf{0.201} & 0.830& 0.916 & 0.988 & \textbf{0.188}&  0.854& 0.947 &  0.986 & \underline{0.192} & 0.889& 0.944 & 0.987 & 0.693 & 0.137 & 0.029 & 0.083\\
  \midrule
  EA only &- & 0.760& 0.901  & 0.980 & -& 0.860& 0.929 & 0.984&-& 0.897 & 0.958& 0.992 &-&-&-&-\\
  PAD only & -&-&-&-&-&-&-&-&-&-&-&-& 0.918&0.063&0.004&0.034\\
  MTL & 0.184 & 0.768 & 0.891 & 0.981 & 0.170 & 0.819 & 0.927 & 0.993 & 0.152 & 0.865 & 0.956 & 0.993 & 0.879 & 0.082 & 0.011 & 0.046  \\
  MTMT \cite{li2020knowledge} & 0.168 & 0.777 & 0.891& 0.979& 0.144 & 0.851 & 0.927 & 0.990& 0.105 & 0.869 & 0.961 & 0.993 & 0.883 & 0.104 & 0.009& 0.057\\
  \rowcolor{Gray}
  EyePAD & \underline{0.162} & 0.718 & 0.852 & 0.976 & \underline{0.128} & 0.757 & 0.891& 0.984 & \underline{0.094} & 0.824 & 0.922 & 0.991 & 0.931 & 0.058 & 0.005 & 0.032\\
  \rowcolor{Gray}
  EyePAD++ & \textbf{0.144} & 0.777& 0.882 & 0.979 & \textbf{0.111} & 0.797 & 0.919 & 0.983  & \textbf{0.082}& 0.878& 0.948 & 0.991&0.926&0.065&0.007&0.036 \\
\bottomrule
\end{tabular}
}
\vspace{-0.3cm}
\caption{\small EA and PAD with Densenet121 trained and evaluated on \textbf{(top)} original, \textbf{(middle)} blurred, \textbf{(bottom)} Noisy data: For user-to-user verification, we report TAR@FAR=$10^{-4},10^{-3},10^{-2}$. For PAD, we report  TDR@FDR=0.002 and APCER, BPCER, HTER. OFRR jointly measures EA and PAD performance on ND-Iris-0405. \textit{EyePAD++ obtains the lowest OFRR}. \textbf{Bold}: Best, \underline{Underlined}: Second best} \label{tab:dnnoblr}
\vspace{-0.4cm}
\end{table*}
\begin{table}[]
\centering
%\scriptsize
\scalebox{0.8}{
\hskip-0.4cm\begin{tabular}{c|cccc|cc}
\toprule
 & \multicolumn{4}{c|}{EA} & \multicolumn{2}{c}{PAD}\\
 \midrule
 & \multicolumn{4}{c|}{1 Query 5 Gallery}&\\
 \midrule
  Method &  OFRR$(\downarrow)$ &$10^{-4}$ &  $10^{-3}$ &$10^{-2}$&   TDR$(\uparrow)$& HTER$(\downarrow)$ \\
  \midrule
  EA only &- & 0.919& 0.983& 0.996&-&-\\
  PAD only &-&-&-&-& 0.962 & 0.026\\
  MTL &  0.113 &  0.815&0.930& 0.977 & 0.921 &0.029\\
  MTMT \cite{li2020knowledge}&  0.068 & 0.891&0.945 & 0.984 & 0.959 & 0.017 \\
  \rowcolor{Gray}
 EyePAD & \underline{0.034} & 0.898 & 0.968 & 0.995 & 0.934 &0.029\\
 \rowcolor{Gray}
  EyePAD++ & \textbf{0.031} & 0.926&  0.976& 0.997 & 0.915&0.031\\
  \midrule
  EA only &- & 0.911 & 0.968& 0.992 &-&- \\
  PAD only &-&-&-&-& 0.801&0.058\\
  MTL & 0.293 & 0.638 & 0.801 & 0.924 & 0.737 & 0.186\\
  MTMT \cite{li2020knowledge} & \underline{0.129} & 0.904 & 0.963 & 0.994 & 0.646 & 0.086 \\
  \rowcolor{Gray}
  EyePAD & 0.132 & 0.902& 0.960& 0.993 & 0.766 & 0.062\\
  \rowcolor{Gray}
  EyePAD++ & \textbf{0.118}&  0.915 & 0.971 & 0.989& 0.655 & 0.067 \\
 \midrule
 EA only &-&0.837& 0.952& 0.992&-&- \\
  PAD only &-&-&-&-& 0.942&0.029\\
  MTL & 0.236& 0.587 & 0.787 & 0.940 & 0.899 &0.045 \\
 MTMT \cite{li2020knowledge}  & 0.114 & 0.824 & 0.916 & 0.975 & 0.894 & 0.034\\
 \rowcolor{Gray}
  EyePAD & \textbf{0.091} & 0.800 & 0.916 & 0.986 &0.914 & 0.033 \\
  \rowcolor{Gray}
  EyePAD++ & \underline{0.093}& 0.840& 0.937& 0.989& 0.887&0.036\\
\bottomrule
\end{tabular}
}
\vspace{-0.3cm}
\caption{\small EA and PAD with HRnet64, trained and evaluated on the \textbf{(top)} original, \textbf{(middle)} blurred, \textbf{(bottom)} noisy (AWGN $\sigma$=3.0)  data: For user-to-user verification, we report  TAR@FAR=$10^{-4},10^{-3},10^{-2}$. For PAD, we report TDR@FDR=0.002 and HTER. \textit{EyePAD++ generally obtains the lowest OFRR}. \textbf{Bold}: Best, \underline{Underlined}: Second best. Results with 1 and 2 gallery pairs are provided in the supplementary material.} \label{tab:hr1q5g}
\vspace{-0.45cm}
\end{table}
\begin{table}[]
\centering
%\scriptsize
\scalebox{0.8}{
\hskip-0.4cm\begin{tabular}{c|cccc|cc}
\toprule
 & \multicolumn{4}{c|}{EA} & \multicolumn{2}{c}{PAD}\\
 \midrule
 & \multicolumn{4}{c|}{1 Query 5 Gallery}&\\
 \midrule
  Method &  OFRR$(\downarrow)$ &$10^{-4}$ &  $10^{-3}$ &$10^{-2}$&   TDR$(\uparrow)$& HTER$(\downarrow)$ \\
  \midrule
  EA only &-&  0.898& 0.952 & 0.995&-&- \\
  PAD only &-&-&-&-&0.925&0.029\\
  MTL & \underline{0.110} & 0.872 &0.933&0.985& 0.884 & 0.039 \\
  MTMT \cite{li2020knowledge} & 0.126 & 0.859 & 0.933 & 0.987& 0.793 &0.042 \\
  \rowcolor{Gray}
 EyePAD & 0.114 & 0.887 &0.947  &0.991 & 0.859 & 0.040\\
 \rowcolor{Gray}
 EyePAD++ & \textbf{0.085 }& 0.901&0.962& 0.990 & 0.883 & 0.032\\
  \midrule
   EA only  &- & 0.846 & 0.921 & 0.989 & - & - \\
  PAD only &-&-&-&-& 0.581 & 0.117\\
  MTL & 0.483 & 0.744&0.855&0.942 & 0.556  & 0.128 \\
  MTMT \cite{li2020knowledge} & 0.464 & 0.769 & 0.913 & 0.963 & 0.502& 0.137  \\
  \rowcolor{Gray}
  EyePAD & \underline{0.447}& 0.711&0.866& 0.956 & 0.589 & 0.121\\
  \rowcolor{Gray}
  EyePAD++ & \textbf{0.332} & 0.861 &  0.938 & 0.983 & 0.552 & 0.133\\
 \midrule
 EA only  & - & 0.817 & 0.944 & 0.985 &-&-\\
  PAD only &-&-&-&-& 0.831 & 0.041\\
  MTL & 0.173 & 0.761 &0.908&0.974& 0.762 &  0.064\\
  MTMT \cite{li2020knowledge}  &\underline{0.162}& 0.777&0.906&0.977& 0.730 & 0.059\\
  \rowcolor{Gray}
  EyePAD & 0.209 & 0.801 & 0.927& 0.980 & 0.712 & 0.065\\
  \rowcolor{Gray}
  EyePAD++  &\textbf{0.137} & 0.811&0.912& 0.990& 0.730 &  0.080\\
\bottomrule
\end{tabular}
}
\vspace{-0.3cm}
\caption{\small EA and PAD with MobilenetV3, trained and evaluated on the \textbf{(top)} original, \textbf{(middle)} blurred, \textbf{(bottom)} noisy (AWGN $\sigma$=3.0)  data: For user-to-user verification, we report  TAR@FAR=$10^{-4},10^{-3},10^{-2}$. For PAD, we report TDR@FDR=0.002 and HTER. \textit{EyePAD++ generally obtains the lowest OFRR}. \textbf{Bold}: Best, \underline{Underlined}: Second best. Results with 1 and 2 gallery pairs are provided in the supplementary material.} \label{tab:mb1q5g}
\vspace{-0.45cm}
\end{table}

\subsection{Metrics for EA and PAD}
Performing the similarity computation (Eq. \ref{eq:sim}) for every possible pairs from $(Q,G)$ and varying the similarity threshold for deciding match/non-match, we compute the ROC curve and report the True Acceptance Rates (TARs) at FAR=$10^{-4}, 10^{-3}, 10^{-2}$. In the biometrics literature \cite{maze2018iarpa, ranjan2019fast, dhar2019measuring}, it is common to use several gallery samples in authentication. But, for authentication on edge devices, the number of gallery samples that can be used depends on the storage capacity of the edge device. Therefore, for evaluating the EA performance of a given model, we use one query left-right pair and $K=1/2/5$ gallery left-right pair(s) for verification.

PAD performance is evaluated with four commonly used metrics: (i) True Detection Rate (TDR) at a False Detection Rate of 0.002, (ii) Attack Presentation Classifier Error Rate (APCER), that is the fraction of spoof samples misclassified as Live, (iii) Bonafide Presentation Classifier Error Rate (BPCER), that is the fraction of live samples misclassified as spoof, (iv) Half Total Error Rate (HTER), the average of APCER and BPCER. Following the protocol in \cite{yambay2017livdet}, we use a threshold of 0.5 for computing APCER and BPCER.

While these metrics gauge either the EA or PAD performance, they cannot jointly evaluate PAD and EA. Hence, we define a new metric in the next subsection. \vspace{-0.3cm}
\subsubsection{Overall False Reject Rate (OFRR)}
We evaluate EA performance on the test EA data (ND-Iris-0405) and the PAD performance on the test subset of CU-LivDet datasets from the 2013, 2015 and 2017 challenges. An ideal metric must measure PAD and EA performance simultaneously on a single dataset. Such a metric must measure: \textit{How often does the model reject true users from accessing the system?} A true user in the EA dataset can be falsely rejected as: (1) `Spoof' by the PAD pipeline, or (2) `Non-match' by the user-to-user verification pipeline.  In this regard, we introduce a new metric called \textbf{O}verall \textbf{F}alse \textbf{R}ejection \textbf{R}ate (OFRR) for true query users in the EA test subset. The steps for computing the OFRR of true users are summarized in Protocol \ref{prot:ofrr}. To determine OFRR, we must first set thresholds for the rates at which PAD misclassifies spoof as live (i.e. Spoof Acceptance Rate or SAR) and EA falsely accepts non-match pairs (FAR). The PAD threshold $t_{pad}$ is computed as the point where SAR=5\% when PAD test data is fed to the model. Similarly, when EA test data is fed to the model, the similarity threshold $t_{auth}$ is computed as the point where the user-to-user verification results in FAR=$10^{-3}$. After computing $t_{pad}$ and $t_{auth}$, we feed the EA test data to the model again. We then compute the number of query users falsely rejected as spoof ($X_{spoof}$) using $t_{pad}$. Here, we reject a query user if at least one of the associated eye images is classified as spoof (Line 12 in Protocol \ref{prot:ofrr}). For those query users classified as live, we verify them for EA against matching gallery users, using our user-to-user verification protocol (Fig \ref{fig:u2u}). We compute the number of query users that are falsely rejected as non-match $X_{auth}$ using $t_{auth}$. Finally, we compute the Overall False Reject Rate (OFRR) as 
\begin{equation}
    \text{OFRR}=\frac{ X_{spoof} + X_{auth}}{|Q|} 
\end{equation}
 $|Q|$ = total number of query users in the EA test dataset, which, in our case, is 150 (Sec. \ref{subsec:datarch}). Ideally, \textit{an authentication system for EA and PAD must have a lower OFRR}.\\ 

The user-to user verification and OFRR protocols are run consecutively, and depend on random samples of left and right images of query and gallery users as shown in Protocols \ref{prot:u2u} (Lines 5,6,7,8) and \ref{prot:ofrr} (Lines 2, 3). Hence, we compute these metrics ten times and report the average.

\subsection{Results}
\textbf{EA and PAD with Densenet backbone:} We perform the EA and PAD experiments using the Densenet121 backbone for the original datasets  and degraded datasets (Table \ref{tab:dnnoblr}). \textit{EyePAD++ obtains the lowest OFRR in most of the the problem settings}. Moreover, EyePAD++ obtains higher user-to-user verification performance than existing multitasking baselines at most FARs. This demonstrates the advantage of the additional distillation step combined with MTL. The PAD performance demonstrated by EyePAD++ is also comparable to that of other multitasking baselines.\\

\textbf{EA and PAD with HRnet64 and MobilenetV3 backbone:} To demonstrate the generalizability of our proposed methods,  we repeat the same experiments with the HRnet64\cite{wang2020deep} backbone (Table \ref{tab:hr1q5g}). However, training a Densenet or HRnet64 model is computationally expensive. So, we also perform the same experiment with the MobilenetV3 \cite{howard2019searching} backbone, that is much more computationally efficient than Densenet (Table \ref{tab:mb1q5g}). More detailed results with 1 or 2 gallery pairs are provided in the supplementary material. Once again we find that \textit{EyePAD++ obtains lower OFRR than MTL and MTMT\cite{li2020knowledge}}. The superiority of EyePAD++ with MobilenetV3 indicates that EyePAD++ can be used for performing EA and PAD on compute engines with low capacity that are available on edge devices. \\%\textbf{EA and PAD with MobilenetV3 backbone: }Our proposed methods: EyePAD and EyePAD++ outperform the MTL baselines when we use a Densenet backbone. However, training a Densenet model is computationally expensive. So, we perform the same experiment with the MobilenetV3 \cite{howard2019searching} backbone, that is much more computationally efficient than Densenet. The results are presented Tables \ref{tab:mbnoblur}, \ref{tab:mbblur5} and \ref{tab:mbnoise3}. \textit{EyePAD++ outperforms the baselines in all of the problem scenarios}, and thus can be used for performing EA and PAD on compute engines with low capacity that are available on edge devices.\\

\begin{figure}
\label{fig:tradeoffsupp}
\centering
\subfloat[Eye-to-eye verification]{\includegraphics[width=0.5\linewidth]{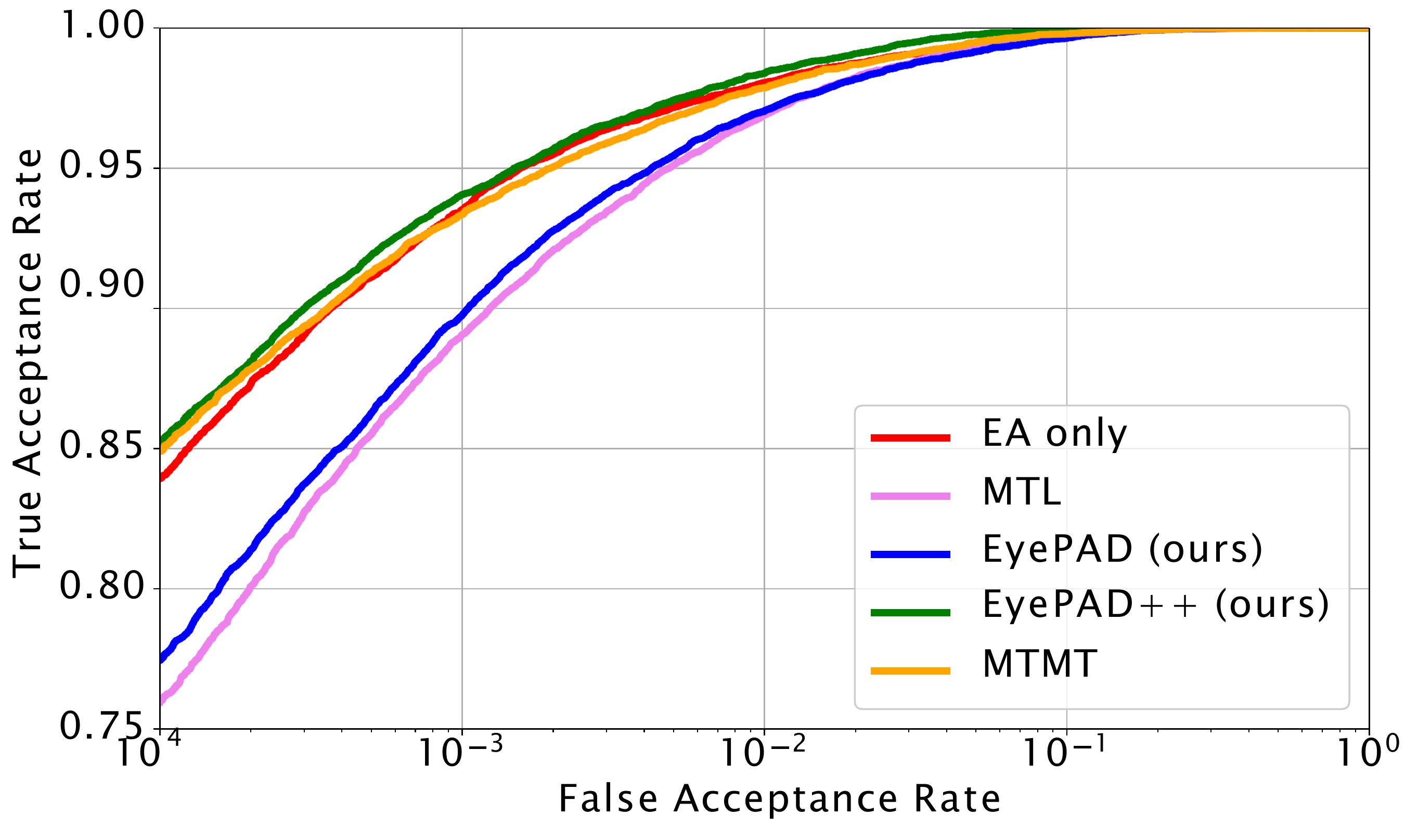}\label{fig:rocir}}
~\subfloat[PAD]{\includegraphics[width=0.5\linewidth]{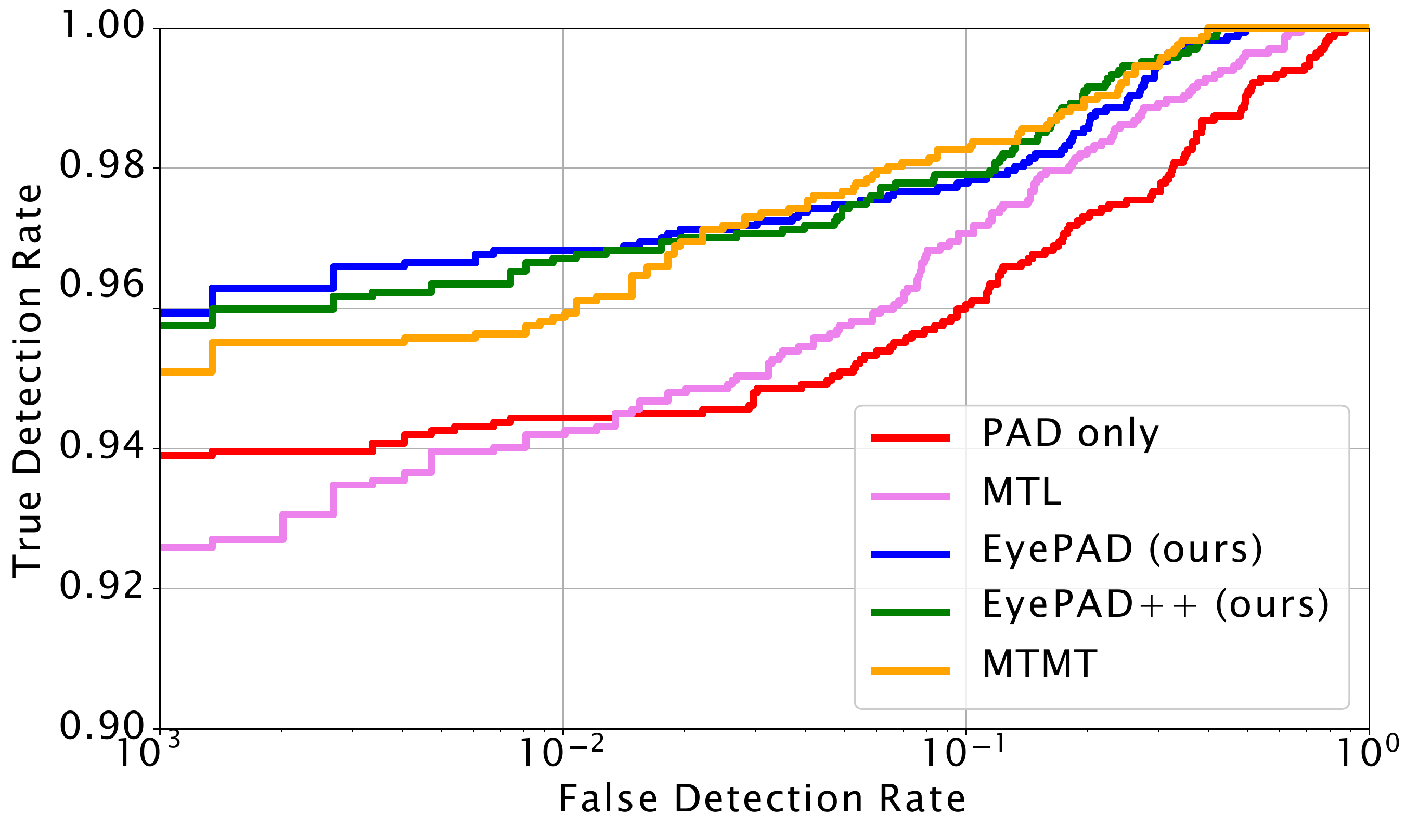}\label{fig:rocpad}}
\vspace{-0.3cm}
\caption{\small ROC curves for (a)EA performance on ND-Iris-0405 (b) PAD performance on CU-LivDet}\vspace{-0.5cm}
\label{fig:roc}
\end{figure}

\textbf{Eye-to-eye verification with PAD: }To compare our proposed methods with current SOTA in PAD and EA, we perform eye-to-eye verification with PAD. Here, for EA, we follow \cite{wang2019toward, zhao2017towards} and use the first 25 left eye images of every user in the ND-Iris-0405 dataset \cite{bowyer2016nd} for training. We use the first 10 right eye images of the users for testing. For evaluating EA, we use the same eye-to-eye verification in \cite{wang2019toward, zhao2017towards}. For PAD, we follow \cite{fang2021iris, sharma2020d} and only use the official train and test split of the CU-LivDet-2017 dataset. We perform this experiment with Densenet121. While training and testing our methods and baselines (Sec. \ref{subsec:baseline}), we exclude pre-processing. Fig. \ref{fig:roc} shows the ROC curves for EA and PAD obtained by all the methods. From Table \ref{tab:i2i}, we infer that \textit{EyePAD and EyePAD++ achieve better PAD performance (i.e. TDR @ FDR=0.002) than the current SOTA PAD algorithms}, without any pre-processing. Moreover, \textit{EyePAD++ achieves higher EA performance (TAR at FAR=$10^{-3}$) than the comparable baselines} (i.e. EA only network, MTL and MTMT). The EA performance for EyePAD and EyePAD++ is comparable to but slightly lower than that of the SOTA \cite{wang2019toward, zhao2017towards}, with a difference of less than 4\%.  We believe that this is difference is due to excluding pre-processing steps for limiting computational cost.\\
\begin{table}[]
\centering
%\scriptsize
\scalebox{0.8}{
\hskip-0.35cm\begin{tabular}{c|cc|ccccc}
\toprule
 & \multicolumn{2}{|c|}{EA} & \multicolumn{4}{c}{PAD}\\
 \midrule
  Method  & TAR$(\uparrow)$ & EER &  TDR$(\uparrow)$ & APCER &BPCER& HTER \\
  \midrule
  IrisCode\textsuperscript{$\dag$} \cite{masek2003recognition}&0.967&1.88&-&-&-&-\\
  Ordinal\textsuperscript{$\dag$} \cite{sun2008ordinal}&0.968&1.74&-&-&-&-\\
  UniNet\textsuperscript{$\dag$} \cite{zhao2017towards}&0.971&1.40&-&-&-&-\\
  DRFnet\textsuperscript{$\dag$} \cite{wang2019toward} & \textbf{0.977} & 1.30&-&-&-&-\\
  Winner of \cite{yambay2017livdet} \textsuperscript{$\dag$} &-&-&-&13.39&0.89&7.10\\
  SpoofNet\textsuperscript{$\dag$} \cite{kimura2020cnn}&-&-&-&33.00&0.00&16.50\\
  Meta-fusion\textsuperscript{$\dag$} \cite{kuehlkamp2018ensemble} &-&- &-&18.66&0.24&9.45\\
  D-net-PAD\textsuperscript{$\dag$} \cite{sharma2020d} & - &-&92.05&5.78&0.94&3.36\\
  PBS \cite{fang2021iris} & - & -& 94.02 & 8.97 & 0.0 & 4.48 \\
  A-PBS \cite{fang2021iris}  & - & -& 92.35 & 6.16 & 0.81 &3.48\\
  \midrule
  EA only&0.936&1.48 &-&-&-&-  \\
  PAD only&- &- &94.02&5.96&0.02&2.99 \\
  MTL &0.891&1.88&92.70&9.03&0.00&4.52 \\
  MTMT \cite{li2020knowledge}& 0.933 & 1.54 & 95.51& 7.54 & 0.00 & 3.77\\
  \rowcolor{Gray}
  EyePAD &0.898&1.89&\textbf{96.29}&5.68&0.00&2.84\\
  \rowcolor{Gray}
  EyePAD++ &0.941&1.30&95.99&7.29&0.00&3.65 \\
 \midrule
\bottomrule
\end{tabular}
}
\vspace{-0.3cm}
\caption{\small Eye-to-eye verification (TAR@FAR=$10^{-3}$ and Equal Error Rate) and PAD performance (TDR@FDR=0.002). \textsuperscript{$\dag$}=Use pre-processing (Segmentation/detection)} \label{tab:i2i}
\vspace{-0.4cm}
\end{table}

\textbf{EyePAD++ v/s MTMT \cite{li2020knowledge}:} Both EyePAD++ and MTMT combine MTL with feature-level KD. However, the student MTL network in MTMT does not inherit the `versatility' through distillation since its teachers are single-task models that are not versatile. On the other hand, EyePAD++ uses distillation from a single versatile teacher ($M_s$), that is proficient in both the tasks. As a result, the student network $M^{*}_s$ in EyePAD++ inherits the versatility of its teacher network $M_s$ through distillation. This enables EyePAD++ to outperform \cite{li2020knowledge} in almost all of problem settings (Tables \ref{tab:dnnoblr},\ref{tab:hr1q5g},\ref{tab:mb1q5g},\ref{tab:i2i}). Thus, for training an MTL network with distillation, we show that \textit{using a single teacher proficient in both the tasks is better than using two teachers proficient in single tasks} in our disjoint multitasking problem.
\section{Conclusion}
In this work, we propose two knowledge distillation-based frameworks: EyePAD and EyePAD++ for joint EA and PAD tasks. For evaluating EA, we present a new user-to-user verification protocol and introduce a new metric to jointly measure user-to-user verification and PAD. Our proposed methods outperform the existing baselines (MTL and MTMT) in most of the problem settings. We evaluate our methods using different network backbones and multiple image quality degradation. Additionally, we evaluate our methods to perform  eye-to-eye verification with PAD (following previous work). Although we do not use any pre-processing, EyePAD and EyePAD++ outperform the SOTA in PAD and obtain eye-to-eye verification performance that is comparable to  SOTA EA algorithms.
\section*{Acknowledgement}
This work was done when the first author was an intern at Meta. This research is partially supported by a MURI from the Army Research Office under the Grant No. W911NF-17-1-0304. This is part of the collaboration between US DOD, UK MOD and UK Engineering and Physical Research Council (EPSRC) under the Multidisciplinary University Research Initiative.
{\small
\bibliographystyle{ieee_fullname}
\bibliography{egbib}
}
\setcounter{section}{0}
\renewcommand{\thesection}{A\arabic{section}}  
\setcounter{table}{0}
\renewcommand{\thetable}{A\arabic{table}}  
\setcounter{figure}{0}
\renewcommand{\thefigure}{A\arabic{figure}}
\setcounter{equation}{0}
\renewcommand{\theequation}{A\arabic{equation}}
\section*{Supplementary material}
In this supplementary material, we provide the following information:\\
\textbf{Section \ref{sec:split}}: Train and test split for user-to-user verification. \\
\textbf{Section \ref{sec:hpeyepad}}: Hyperparameters for EyePAD and EyePAD++. \\
\textbf{Section \ref{sec:detailedresult}}: Detailed results with HRnet and MobilenetV3 backbones.\\
\textbf{Section \ref{sec:ablation}}: Ablation experiments for $\lambda_1$ (EyePAD). \\
\textbf{Section \ref{sec:hpbaseline}}: Hyperparameters for baseline methods.
\section{Train and test splits for ND-Iris-0405 dataset \cite{bowyer2016nd}}
\label{sec:split}
In section 4.2 of the main paper, we mention that we randomly split the users into two subsets: $U_{train}$ (for training) and $U_{test}$ (for evaluation). All the images for users in the training split are used for training. Here, we provide the train and test split to enable researchers reproduce our protocol.\\

\textbf{Train user IDs}: `04200' `04203' `04214' `04233' `04239' `04261' `04265' `04267' `04284'
 `04286' `04288' `04302' `04309' `04313' `04320' `04327' `04336' `04339'
 `04349' `04351' `04361' `04370' `04378' `04379' `04382' `04387' `04394'
 `04395' `04397' `04400' `04407' `04408' `04409' `04418' `04419' `04429'
 `04430' `04434' `04435' `04436' `04440' `04446' `04447' `04453' `04460'
 `04471' `04472' `04475' `04476' `04477' `04479' `04481' `04482' `04485'
 `04495' `04496' `04502' `04504' `04505' `04506' `04511' `04512' `04514'
 `04530' `04535' `04542' `04553' `04560' `04575' `04577' `04578' `04581'
 `04587' `04588' `04589' `04593' `04596' `04597' `04598' `04603' `04605'
 `04609' `04613' `04615' `04622' `04626' `04628' `04629' `04632' `04633'
 `04634' `04644' `04647' `04653' `04670' `04684' `04687' `04691' `04692'
 `04695' `04699' `04701' `04702' `04703' `04712' `04715' `04716' `04720'
 `04721' `04725' `04729' `04734' `04736' `04737' `04738' `04742' `04744'
 `04745' `04747' `04748' `04751' `04756' `04757' `04758' `04763' `04765'
 `04768' `04772' `04773' `04774' `04776' `04777' `04778' `04782' `04783'
 `04785' `04787' `04790' `04792' `04794' `04797' `04801' `04802' `04803'
 `04813' `04815' `04816' `04818' `04831' `04832' `04839' `04840' `04841'
 `04843' `04846' `04847' `04850' `04854' `04857' `04858' `04859' `04861'
 `04863' `04864' `04866' `04867' `04869' `04870' `04871' `04872' `04873'
 `04876' `04877' `04878' `04879' `04880' `04882' `04883' `04884' `04886'
 `04888' `04890' `04891' `04892' `04894' `04897' `04898' `04899' `04901'
 `04905' `04908' `04909' `04910' `04911' `04912' `04914' `04915' `04919'
 `04920' `04922' `04923' `04928' `04930' `04931' `04932' `04934'\\

\textbf{Test user IDs}:`02463' `04201' `04202' `04213' `04217' `04221' `04225' `04273' `04285'
 `04297' `04300' `04301' `04311' `04312' `04314' `04319' `04322' `04324'
 `04334' `04338' `04341' `04343' `04344' `04347' `04350' `04372' `04385'
 `04388' `04404' `04423' `04427' `04444' `04449' `04451' `04456' `04459'
 `04461' `04463' `04470' `04473' `04488' `04493' `04507' `04509' `04513'
 `04519' `04531' `04537' `04556' `04557' `04569' `04580' `04585' `04595'
 `04600' `04612' `04621' `04631' `04641' `04662' `04664' `04667' `04673'
 `04675' `04681' `04682' `04683' `04689' `04693' `04697' `04705' `04708'
 `04709' `04711' `04714' `04719' `04722' `04724' `04726' `04727' `04728'
 `04730' `04731' `04732' `04733' `04743' `04746' `04749' `04754' `04760'
 `04762' `04767' `04770' `04775' `04784' `04786' `04796' `04798' `04806'
 `04810' `04811' `04812' `04821' `04822' `04823' `04827' `04829' `04830'
 `04833' `04838' `04842' `04848' `04849' `04851' `04853' `04855' `04856'
 `04860' `04862' `04865' `04868' `04874' `04875' `04881' `04885' `04887'
 `04889' `04893' `04895' `04896' `04900' `04902' `04903' `04904' `04906'
 `04907' `04913' `04916' `04917' `04918' `04921' `04924' `04925' `04926'
 `04927' `04929' `04933' `04935' `04936' \\

It is also mentioned in the main paper that the images for the test users are then randomly divided into query and gallery sets. We provide the images in the query and gallery sets in \texttt{query\_set.txt} and \texttt{gallery\_set.txt}, respectively. These files are provided \href{https://drive.google.com/drive/folders/1iTPFq87MknAoQxPz4osmCKGKNGhZmX6V?usp=sharing}{here}. We also provide a readme file (\texttt{README.txt}) for the readers' convenience, wherein we provide information about the left/right labels.
\begin{table*}[]
\centering
%\scriptsize
\scalebox{0.75}{
\hskip-0.7cm\begin{tabular}{c|cccc|cccc|cccc|cccc}
\toprule
 & \multicolumn{12}{c|}{User-to-user verification results on ND-Iris-0405 (EA)} & \multicolumn{4}{c}{PAD results on CU-LivDet}\\
 \midrule
 & \multicolumn{4}{c|}{1 Query 1 Gallery}&  \multicolumn{4}{c|}{1 Query 2 Gallery}&  \multicolumn{4}{c|}{1 Query 5 Gallery}&\\
 \midrule
  Method & OFRR$(\downarrow)$ & $10^{-4}$ &  $10^{-3}$&$10^{-2}$&  OFRR$(\downarrow)$ &$10^{-4}$ &  $10^{-3}$& $10^{-2}$&  OFRR$(\downarrow)$ &$10^{-4}$ &  $10^{-3}$ &$10^{-2}$&   TDR$(\uparrow)$ & APCER &BPCER& HTER$(\downarrow)$ \\
  EA only & -& 0.861& 0.950& 0.990&- & 0.875& 0.961& 0.994 &- & 0.919& 0.983& 0.996&-&-&-&-\\
  PAD only & -&-&-&-&-&-&-&-&-&-&-&-& 0.962 & 0.051& 0.00 & 0.026\\
  MTL & 0.209& 0.682 & 0.819 &0.949 & 0.156 & 0.734& 0.866 &0.962&  0.113 &  0.815&0.930& 0.977 & 0.921 & 0.053&0.005&0.029\\
  MTMT \cite{li2020knowledge}& 0.132 & 0.777 & 0.881 &0.970 & 0.091 & 0.840 & 0.917&0.971 & 0.068 & 0.891&0.945 & 0.984 & 0.959 & 0.033& 0.001& 0.017 \\
  \rowcolor{Gray}
 EyePAD & \underline{0.094}& 0.804 & 0.909&  0.985 &  \underline{0.060} & 0.864 &0.942&0.993& \underline{0.034} & 0.898 & 0.968 & 0.995 & 0.934 & 0.044& 0.014&0.029\\
 \rowcolor{Gray}
  EyePAD++ & \textbf{0.062}& 0.865& 0.942 & 0.996 & \textbf{0.048}& 0.898& 0.959& 0.993 & \textbf{0.031} & 0.926&  0.976& 0.997 & 0.915&0.041&0.021&0.031\\
 \midrule
 EA only & -& 0.829 & 0.913 & 0.978&- & 0.838& 0.944  &0.988& - & 0.911 & 0.968& 0.992 &-&-&-&- \\
  PAD only& -&-&-&-&-&-&-&-&-&-&-&-& 0.801&0.089&0.026&0.058\\
  MTL & 0.430 & 0.503 & 0.650 & 0.851 & 0.377 & 0.509 & 0.719 & 0.868 & 0.293 & 0.638 & 0.801 & 0.924 & 0.737 & 0.040 & 0.331 & 0.186\\
  MTMT \cite{li2020knowledge}& 0.188 & 0.768 & 0.898 & 0.973 & 0.165 & 0.846 & 0.932 & 0.990 & \underline{0.129} & 0.904 & 0.963 & 0.994 & 0.646 & 0.125 & 0.046 & 0.086 \\
  \rowcolor{Gray}
  EyePAD & \underline{0.180} & 0.805& 0.897 & 0.970& \textbf{0.137} & 0.877  &0.932& 0.989& 0.132 & 0.902& 0.960& 0.993 & 0.766 & 0.115 & 0.008 & 0.062\\
  \rowcolor{Gray}
  EyePAD++ &  \textbf{0.174}& 0.830 & 0.911 & 0.983&\underline{0.158} & 0.869& 0.950& 0.988 & \textbf{0.118}&  0.915 & 0.971 & 0.989& 0.655 & 0.109 & 0.025& 0.067 \\
  \midrule
  EA only &-&0.733 & 0.893 & 0.984 &-&0.749& 0.924& 0.990&-&0.837& 0.952& 0.992&-&-&-&- \\
  PAD only &- &-&-&-&-&-&-&-&-&-&-&-& 0.942&0.049&0.008&0.029\\
  MTL & 0.338 & 0.451& 0.678 & 0.856 & 0.279 & 0.481& 0.737 & 0.908 & 0.236& 0.587 & 0.787 & 0.940 & 0.899 & 0.085 & 0.004 &0.045 \\
  MTMT \cite{li2020knowledge} & 0.184 & 0.675 & 0.846 & 0.944& 0.168 & 0.721 & 0.862 & 0.961 & 0.114 & 0.824 & 0.916 & 0.975 & 0.894 & 0.048 & 0.020 & 0.034\\
  \rowcolor{Gray}
  EyePAD & \underline{0.161} & 0.656& 0.848 &  0.964 & \underline{0.125} & 0.718 & 0.886 & 0.984 & \textbf{0.091} & 0.800 & 0.916 & 0.986 &0.914 & 0.060 & 0.006 & 0.033 \\
  \rowcolor{Gray}
  EyePAD++ & \textbf{0.157}& 0.729&0.869 &0.976& \textbf{0.114}&0.781&0.916& 0.988 & \underline{0.093}& 0.840& 0.937& 0.989& 0.887&0.061&0.010&0.036\\
\bottomrule
\end{tabular}
}
\vspace{-0.3cm}
\caption{\small EA and PAD with HRnet64 trained and evaluated on \textbf{(top)} original, \textbf{(middle)} blurred, \textbf{(bottom)} Noisy data: For user-to-user verification, we report TAR@FAR=$10^{-4},10^{-3},10^{-2}$. For PAD, we report  TDR@FDR=0.002 and APCER, BPCER, HTER. OFRR jointly measures EA and PAD performance on ND-Iris-0405. \textit{EyePAD++ obtains the lowest OFRR in most scenarios}. \textbf{Bold}: Best, \underline{Underlined}: Second best.} \label{tab:hrall}
\vspace{-0.2cm}
\end{table*}
\begin{table*}[]
\centering
%\scriptsize
\scalebox{0.75}{
\hskip-0.7cm\begin{tabular}{c|cccc|cccc|cccc|cccc}
\toprule
 & \multicolumn{12}{c|}{User-to-user verification results on ND-Iris-0405 (EA)} & \multicolumn{4}{c}{PAD results on CU-LivDet}\\
 \midrule
 & \multicolumn{4}{c|}{1 Query 1 Gallery}&  \multicolumn{4}{c|}{1 Query 2 Gallery}&  \multicolumn{4}{c|}{1 Query 5 Gallery}&\\
 \midrule
  Method & OFRR$(\downarrow)$ & $10^{-4}$ &  $10^{-3}$&$10^{-2}$&  OFRR$(\downarrow)$ &$10^{-4}$ &  $10^{-3}$& $10^{-2}$&  OFRR$(\downarrow)$ &$10^{-4}$ &  $10^{-3}$ &$10^{-2}$&   TDR$(\uparrow)$ & APCER &BPCER& HTER$(\downarrow)$ \\
  EA only &- &0.824 &0.923 &0.989& - &  0.863& 0.943 &0.987 &-&  0.898& 0.952 & 0.995&-&-&-&- \\
  PAD only &-&-&-&-&-&-&-&-&-&-&-&-&0.925&0.048&0.010&0.029\\
  MTL & 0.180 & 0.739 & 0.856 &0.958 & \underline{0.142}& 0.792 &0.895 &0.970 & \underline{0.110} & 0.872 &0.933&0.985& 0.884 & 0.025 & 0.053 & 0.039 \\
  MTMT \cite{li2020knowledge} & 0.193 & 0.751 & 0.871 & 0.973 & 0.144 & 0.817 & 0.917 & 0.977 & 0.126 & 0.859 & 0.933 & 0.987& 0.793 &0.073&0.011 &0.042 \\
  \rowcolor{Gray}
 EyePAD & \underline{0.160} & 0.769 & 0.893& 0.972 & \underline{0.142} & 0.838& 0.919& 0.985 & 0.114 & 0.887 &0.947  &0.991 & 0.859 & 0.046& 0.034 & 0.040\\
 \rowcolor{Gray}
 EyePAD++ &  \textbf{0.140} & 0.819& 0.904&0.981 & \textbf{0.117} & 0.863 &0.934 &0.992& \textbf{0.085} & 0.901&0.962& 0.990 & 0.883 & 0.041& 0.022& 0.032\\
 \midrule
 EA only & - & 0.889&  0.953 & 0.993 & - & 0.853 &0.929& 0.992 &- & 0.846 & 0.921 & 0.989 & - &-& -& - \\
  PAD only & -&-&-&-&-&-&-&-&-&-&-&-& 0.581& 0.155&  0.078 & 0.117\\
  MTL & 0.564 & 0.585& 0.734 & 0.871 & 0.533 & 0.619 & 0.798 & 0.927& 0.483 & 0.744&0.855&0.942 & 0.556 & 0.133 & 0.122 & 0.128 \\
  MTMT \cite{li2020knowledge} &\underline{0.524} & 0.642 & 0.819 & 0.939& \underline{0.477} & 0.726 & 0.905 & 0.965& 0.464 & 0.769 & 0.913 & 0.963 & 0.502& 0.227& 0.046 & 0.137  \\
  \rowcolor{Gray}
  EyePAD & 0.562& 0.537& 0.715& 0.915 & 0.486& 0.618 & 0.810 & 0.952 & \underline{0.447}& 0.711&0.866& 0.956 & 0.589 & 0.144 & 0.097 & 0.121\\
  \rowcolor{Gray}
  EyePAD++ & \textbf{0.385}& 0.768 & 0.874 & 0.956 & \textbf{0.372} & 0.792& 0.913 &0.970 &  \textbf{0.332} & 0.861 &  0.938 & 0.983 & 0.552 & 0.092 &  0.173 & 0.133\\
  \midrule
  EA only & - & 0.756&0.872 &0.977 &-& 0.795 &0.912 &0.982 & - & 0.817 & 0.944 & 0.985 &-&-&-&-\\
  PAD only& -&-&-&-&-&-&-&-&-&-&-&-& 0.831 & 0.079 &0.003 & 0.041\\
  MTL & 0.263 & 0.643&0.799& 0.949& 0.217 & 0.682& 0.870 &0.976& 0.173 & 0.761 &0.908&0.974& 0.762 & 0.062 & 0.065 &  0.064\\
  MTMT \cite{li2020knowledge} & 0.285 & 0.549 & 0.766&0.928 & \underline{0.185} & 0.705 & 0.868 & 0.968 &\underline{0.162}& 0.777&0.906&0.977& 0.730& 0.049 & 0.069& 0.059\\
  \rowcolor{Gray}
  EyePAD & \underline{0.262} & 0.722&0.847& 0.958&  0.227 & 0.779& 0.880& 0.974  & 0.209 & 0.801 & 0.927& 0.980 & 0.712& 0.083 & 0.047 & 0.065\\
  \rowcolor{Gray}
  EyePAD++ & \textbf{0.254} & 0.660& 0.786 &0.947 & \textbf{0.180} &0.733&0.868 &0.975 &\textbf{0.137} & 0.811&0.912& 0.990& 0.730 & 0.033 & 0.126 &  0.080\\
\bottomrule
\end{tabular}
}
\vspace{-0.3cm}
\caption{\small EA and PAD with MobilenetV3 trained and evaluated on \textbf{(top)} original, \textbf{(middle)} blurred, \textbf{(bottom)} Noisy data: For user-to-user verification, we report TAR@FAR=$10^{-4},10^{-3},10^{-2}$. For PAD, we report  TDR@FDR=0.002 and APCER, BPCER, HTER. OFRR jointly measures EA and PAD performance on ND-Iris-0405. \textit{EyePAD++ obtains the lowest OFRR}. \textbf{Bold}: Best, \underline{Underlined}: Second best.} \label{tab:mball}
\vspace{-0.2cm}
\end{table*}
\begin{table}[]
\centering
%\scriptsize
\scalebox{0.75}{
\hskip-0.4cm\begin{tabular}{ccccccccc}
\toprule
 \midrule
  Backbone & Dataset & $\lambda_1$ & $\lambda_2$ & Optimizer & LR & $\gamma$ & Decay after \\
  \midrule
  Densenet121 & Original & 2.0 & 0.75 & Adam & $10^{-4}$ & 0.5 & 12 \\
  Densenet121 & Blurred & 1.0 & 2.0 & Adam & $10^{-4}$ & 0.5 & 12\\
  Densenet121 & Noisy & 1.0 & 2.0 & Adam & $10^{-4}$ & 0.5 & 12\\
  \midrule
  HRnet64 & Original & 2.0 & 2.0 & Adam & $10^{-4}$ & 0.5 & 12\\
  HRnet64 & Blurred &5.0 & 2.0 & Adam & $10^{-4}$ & 0.5 & 12\\
  HRnet64 & Noisy & 2.0 & 5.0 & Adam & $10^{-4}$ & 0.5 & 12\\
  \midrule
  MobilenetV3 & Original &1.0&0.75&SGD&$10^{-1}$&0.1&15\\
  MobilenetV3 & Blurred &1.0&0.75&SGD&$10^{-1}$&0.1&15\\
  MobilenetV3 & Noisy &5.0&2.0&SGD&$10^{-1}$&0.1&15\\
  
 \midrule
\bottomrule
\end{tabular}
}
\vspace{-0.3cm}
\caption{\small Hyperparameter information for EyePAD and EyePAD++.} \label{tab:hpeyepad}
\vspace{-0.2cm}
\end{table}

\begin{table}[]
\centering
%\scriptsize
\scalebox{0.75}{
\hskip-0.6cm\begin{tabular}{ccccccccc}
\toprule
 \midrule
  Backbone & Dataset & $\lambda_{auth}$ & $\lambda_{pad}$ & Optimizer & LR & $\gamma$ & Decay after \\
  \midrule
  Densenet121 & Original & 1.0 & 1.0 & Adam & $10^{-4}$ & 0.5 & 12 \\
  Densenet121 & Blurred & 0.75 & 0.75 & Adam & $10^{-4}$ & 0.5 & 12\\
  Densenet121 & Noisy & 0.5 & 0.75 & Adam & $10^{-4}$ & 0.5 & 12\\
  \midrule
  HRnet64 & Original & 1.0 & 1.0 & Adam & $10^{-4}$ & 0.5 & 12\\
  HRnet64 & Blurred &0.75 & 0.75 & Adam & $10^{-4}$ & 0.5 & 12\\
  HRnet64 & Noisy & 2.0 & 2.0 & Adam & $10^{-4}$ & 0.5 & 12\\
  \midrule
  MobilenetV3 & Original &1.0&2.0&SGD&$10^{-1}$&0.1&15\\
  MobilenetV3 & Blurred &1.0&1.0&SGD&$10^{-1}$&0.1&15\\
  MobilenetV3 & Noisy &0.1&0.1&SGD&$10^{-1}$&0.1&15\\
  
 \midrule
\bottomrule
\end{tabular}
}
\vspace{-0.3cm}
\caption{\small Hyperparameter information for MTMT \cite{li2020knowledge}.} \label{tab:hpmtmt}
\vspace{-0.2cm}
\end{table}
% \begin{figure*}
% \centering
% \subfloat[MTL]{\includegraphics[width=0.25\linewidth]{images/hist_mt.pdf}\label{fig:histmt}}
% \subfloat[MTMT \cite{li2020knowledge}]{\includegraphics[width=0.25\linewidth]{images/hist_mtmt.pdf}\label{fig:histmtmt}}
% ~\subfloat[EyePAD]{\includegraphics[width=0.25\linewidth]{images/hist_apad.pdf}\label{fig:histapad}}
% ~\subfloat[EyePAD++]{\includegraphics[width=0.25\linewidth]{images/hist_apad++.pdf}\label{fig:histapad++}}
% \vspace{-0.3cm}
% \caption{\small Similarity score distribution for matches and non-matches obtained using (a) MTL baseline, (b) EyePAD (c) EyePAD++. Lower overlap between match and non-match distribution is better.\textit{ EyePAD++ demonstrates lowest overlap.}}
% \label{fig:histall}
% \end{figure*}
\begin{figure*}
{
\centering
\subfloat[1 query 1 gallery]{\includegraphics[width=0.333\linewidth]{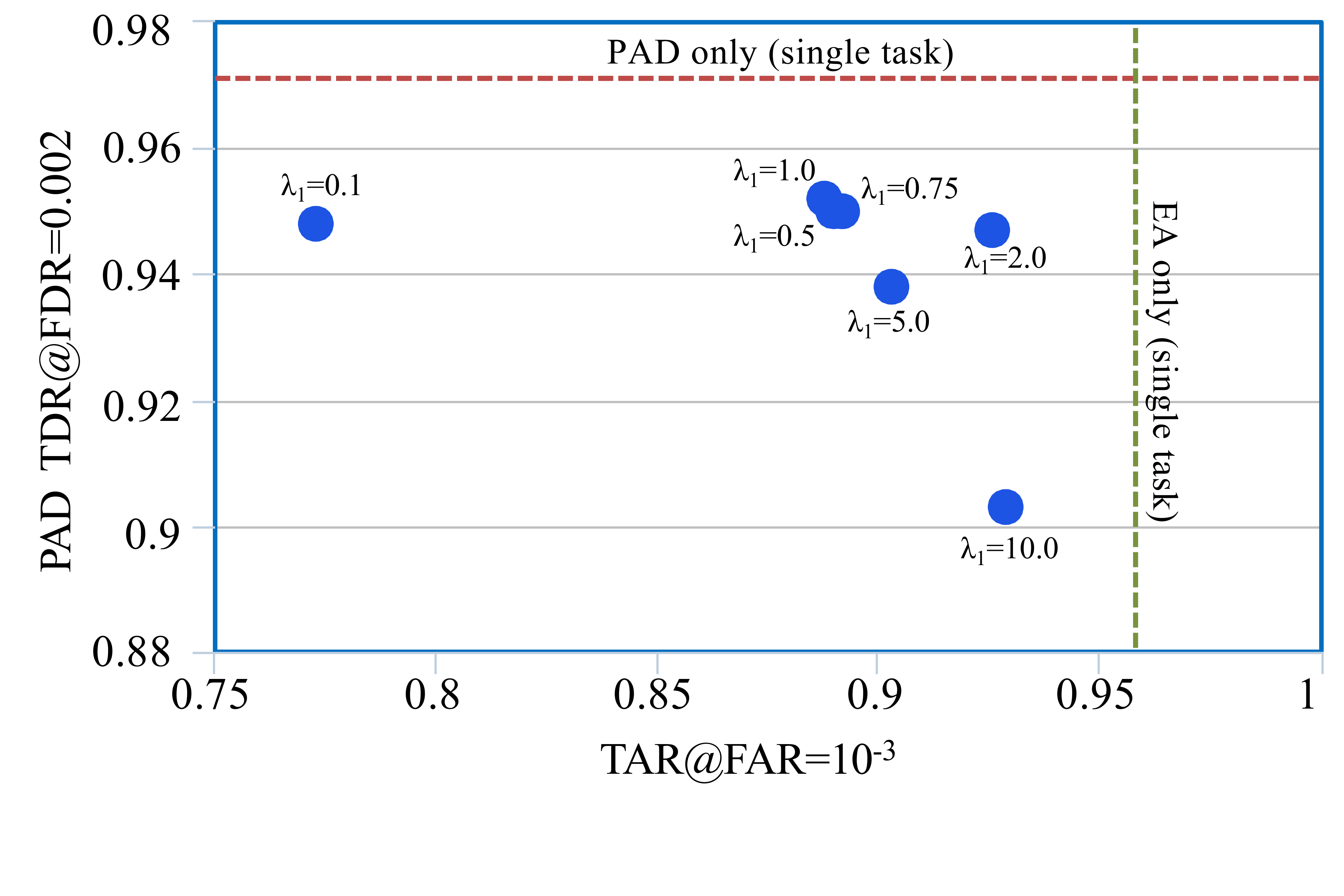}\label{fig:ablation1q1g}}
~\subfloat[1 query 2 gallery]{\includegraphics[width=0.333\linewidth]{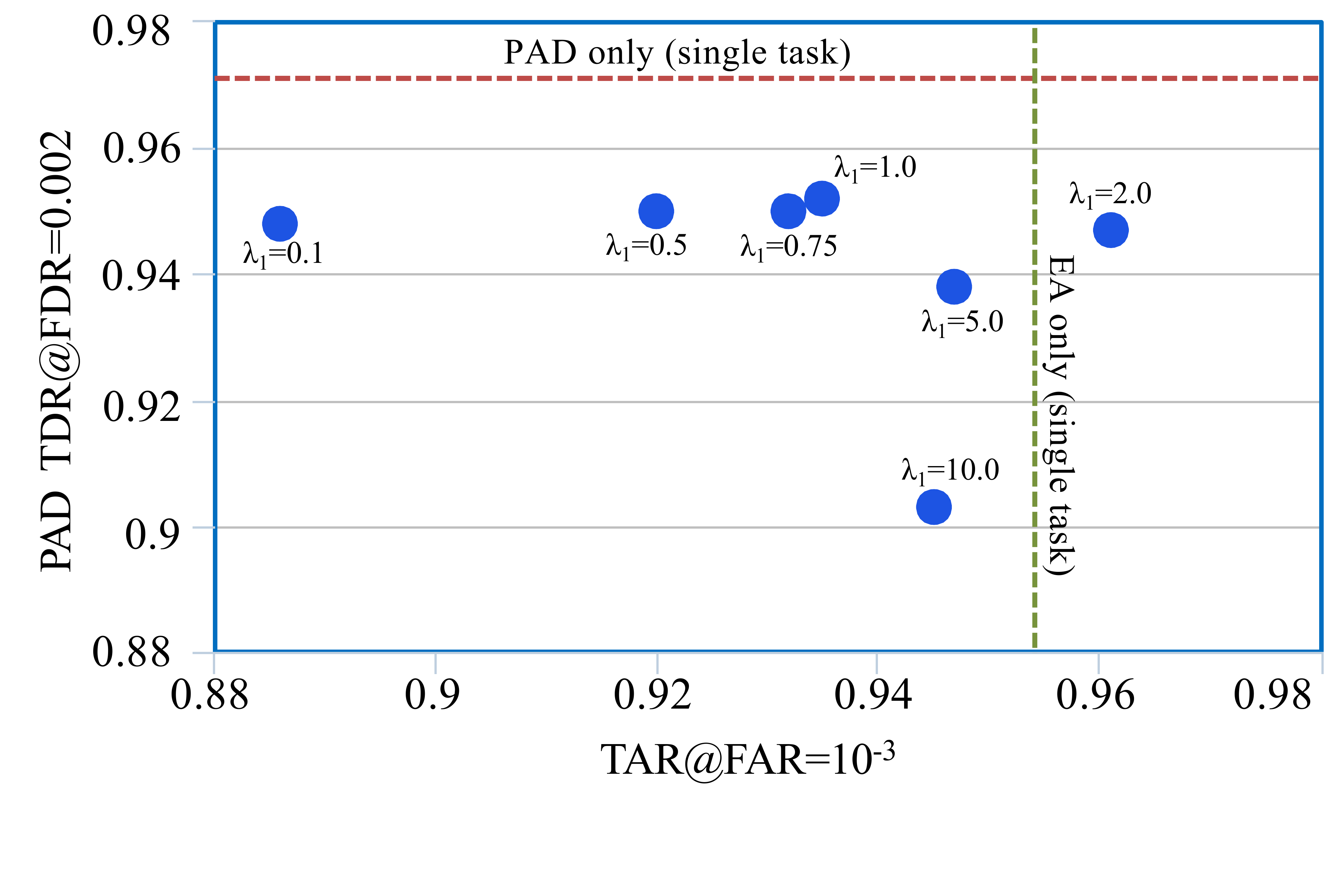}\label{fig:ablation1q2g}}
~\subfloat[1 query 5 gallery]{\includegraphics[width=0.333\linewidth]{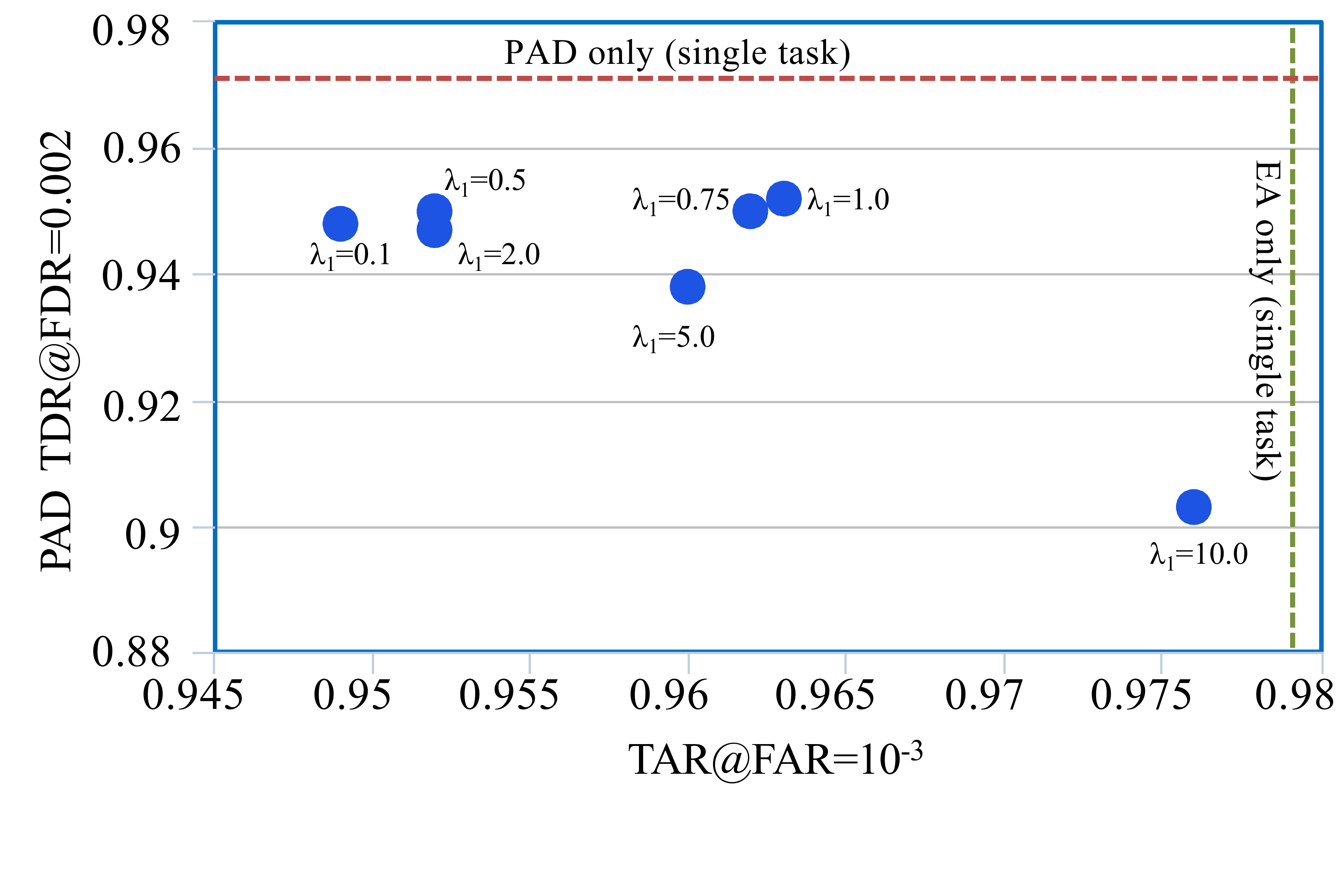}\label{fig:ablation1q5g}}
\vspace{-0.3cm}
\caption{\small PAD performance (TDR@FDR=0.002) v/s User-to-user verification performance (TAR@FAR=$10^{-3}$) obtained by the EyePAD student network, for different values of $\lambda_1$.}\label{fig:ablation}
}
\end{figure*}
\section{Training details for EyePAD, EyePAD++}
\label{sec:hpeyepad}
We train all the models in our work with a batch size of 64 in our experiments, for 100 epochs. We use data augmentation such as random horizontal flip, random rotation (30 degrees) and random jitter. The detailed hyperparameter information for training models for user-to-user verification with PAD is provided in Table \ref{tab:hpeyepad}.

While training Densenet121 network for eye-to-eye verification with PAD, we use $\lambda_1=2.0$ (for EyePAD) and $\lambda_2=2.0$ (for EyePAD++). All the other parameters used in this experiment are same as those mentioned in the first row of Table \ref{tab:hpeyepad}. 
\section{Detailed results}
\label{sec:detailedresult}
%\begin{figure}
%\label{fig:tradeoffsupp}
%\centering
%\subfloat[Densnet121]{\includegraphics[width=0.5\linewidth]{images/dns_apad_1e-4.png}\label{fig:dntartdr}}
%~\subfloat[HRnet64]{\includegraphics[width=0.5\linewidth]{images/hrnet_apad_1e-4.png}\label{fig:hrtartdr}}
%\vspace{-0.3cm}
%\caption{\small PAD TDR@FDR=0.002 v/s IR TAR @ FAR=$10^{-4}$ for networks trained on the original (clean) datasets}
%\end{figure}
\subsection{EA and PAD with HRnet64 }
In Table \ref{tab:hrall}, we provide the full version of Table 4 from the main paper, where we present results with HRnet64 \cite{wang2020deep}. Here, we report the user-to-user verification performance with $K=1,2,5$ gallery pairs. In most of the scenarios, \textit{EyePAD++ outperforms the existing baselines in terms of the OFRR score}. 
\subsection{EA and PAD with MobilenetV3}
In Table \ref{tab:mball}, we provide the full version of Table 5 from the main paper, where we present results with MobilenetV3 \cite{howard2019searching}. Here, we report the user-to-user verification performance with $K=1,2,5$ gallery pairs. \textit{EyePAD++ outperforms the existing baselines in terms of the OFRR score}, in all the problem settings.
\section{Ablation study: Effect of $\lambda_1$ in EyePAD}
\label{sec:ablation}
The hyperparameter $\lambda_1$ is used to weight the feature-level distillation loss $L_{dis}$ (Eq 2 of the main paper). $L_{dis}$ is used to preserve the EA information, while student $M_s$ (initialized with $M_t$) is trained for PAD. Using Densenet121 and the original (clean) EA and PAD datasets, we analyze the effect of $\lambda_1$ on user-to-user verification performance and PAD performance. We perform experiments with $\lambda_1=[0.1,0.5,0.75,1.0,2.0,5.0,10.0]$ and present the corresponding results in Fig. \ref{fig:ablation}. We find that in general, when $\lambda_1$ increases, the user to user verification performance improves and the PAD performance gets degraded. This is expected because a higher value for $\lambda_1$ enforces $M_s$ to preserve authentication and restrics it from learning PAD-specific features. However, we do not find any such trend with respect to parameter $\lambda_2$.
\section{Training details for baselines}
\label{sec:hpbaseline}
For training the MTL baseline for user-to-user or eye-to-eye verification with PAD, we use the same parameter values mentioned in Table \ref{tab:hpeyepad}, except for $\lambda_1, \lambda_2$. The hyperparameters used for training MTMT \cite{li2020knowledge} are provided in Table \ref{tab:hpmtmt}. While training MTMT \cite{li2020knowledge} for eye-to-eye verification with PAD (using Densenet121 and the original dataset), we use $\lambda_{auth}=1.0$ and $\lambda_{pad}=1.0$. The rest of the hyperparameters are same as those mentioned in the first row of Table \ref{tab:hpmtmt}.
\end{document}